\documentclass[11pt]{article}

\ifdefined\pdfinfoomitdate\pdfinfoomitdate=1\fi
\ifdefined\pdftrailerid\pdftrailerid{}\fi
\ifdefined\pdfsuppressptexinfo\pdfsuppressptexinfo=-1\fi
\usepackage[a4paper,margin=2.5cm]{geometry}
\usepackage[T1]{fontenc}
\usepackage[dvipsnames]{xcolor}
\usepackage{newpxtext}
\usepackage{microtype}
\usepackage{setspace}
\usepackage{etoolbox}
\patchcmd{\section}
  {\normalfont\Large\bfseries}
  {\normalfont\Large\sffamily\bfseries}
  {}{\PackageError{manuscript}{Could not set section heading font}{}}
\patchcmd{\subsection}
  {\normalfont\large\bfseries}
  {\normalfont\large\sffamily\bfseries}
  {}{\PackageError{manuscript}{Could not set subsection heading font}{}}

\makeatletter
\patchcmd{\subsection}
  {-3.25ex}
  {-2.5ex}
  {}{\PackageError{manuscript}{Could not set subsection heading spacing before}{}}
\patchcmd{\subsection}
  {1.5ex \@plus .2ex}
  {0.5ex \@plus .2ex}
  {}{\PackageError{manuscript}{Could not set subsection heading spacing}{}}

\renewcommand{\paragraph}{\@startsection{paragraph}{4}{\z@}{0.8ex \@plus 0.2ex \@minus 0.1ex}{-0.6em}{\normalfont\normalsize\itshape}}
\makeatother

\usepackage{authblk}

\renewcommand{\Affilfont}{\footnotesize}
\DeclareRobustCommand{\AffiliationBlock}[2]{\parbox[t]{0.96\textwidth}{\Affilfont\setstretch{1}\raggedright
    \setlength{\parindent}{0pt}\setlength{\parskip}{0pt}\hangindent=1.4em\hangafter=1
    \noindent\makebox[1.4em][l]{\textsuperscript{#1}}#2\strut\par}}
\usepackage{letltxmacro}
\LetLtxMacro{\OriginalAffil}{\affil}
\renewcommand{\affil}[2][]{\OriginalAffil[ ]{\AffiliationBlock{#1}{#2}}}
\makeatletter
\renewcommand{\AB@affilsepx}{\protect\\[2.5pt]\protect\Affilfont}
\makeatother

\makeatletter
\patchcmd{\@maketitle}
  {\LARGE \@title}
  {\LARGE\rmfamily\mdseries \@title}
  {}{\PackageError{manuscript}{Could not set article title font}{}}
\patchcmd{\@maketitle}
  {\vskip 1em{\large \@date}}
  {\ifx\@date\@empty\else\vskip 1em{\large \@date}\fi}
  {}{\PackageError{manuscript}{Could not remove empty date spacing}{}}
\BeforeBeginEnvironment{abstract}{\vspace*{-2\baselineskip}}
\patchcmd{\abstract}
  {\bfseries \abstractname}
  {\sffamily\bfseries \abstractname}
  {}{\PackageError{manuscript}{Could not set abstract heading font}{}}
\makeatother

\usepackage{amsmath,mathtools}
\usepackage{newpxmath}
\usepackage{bm}

\usepackage{graphicx}
\usepackage{booktabs,array,tabularx,longtable,multirow,threeparttable}
\usepackage{caption,subcaption}
\DeclareCaptionType{methodfigure}[Methods Fig.][List of Methods Figures]
\usepackage[section]{placeins}

\usepackage{enumitem}
\usepackage{xspace}

\usepackage[numbers,sort&compress]{natbib}

\definecolor{linkblue}{HTML}{1F4E79}
\usepackage[
  pagebackref=false,
  breaklinks=true,
  colorlinks=true,
  citecolor=linkblue,
  linkcolor=linkblue,
  urlcolor=linkblue
]{hyperref}
\hypersetup{pdfcreator={},pdfproducer={}}
\usepackage[capitalise,noabbrev]{cleveref}
\crefname{methodfigure}{Methods Fig.}{Methods Figs.}
\Crefname{methodfigure}{Methods Fig.}{Methods Figs.}
\crefname{figure}{Fig.}{Figs.}
\Crefname{figure}{Figure}{Figures}
\crefname{table}{Table}{Tables}
\Crefname{table}{Table}{Tables}
\crefname{equation}{Eq.}{Eqs.}
\Crefname{equation}{Equation}{Equations}

\crefname{suppfigure}{Supplementary Fig.}{Supplementary Figs.}
\Crefname{suppfigure}{Supplementary Figure}{Supplementary Figures}
\crefname{supptable}{Supplementary Table}{Supplementary Tables}
\Crefname{supptable}{Supplementary Table}{Supplementary Tables}

\newcommand{\simulationexecutionref}{\cref{mfig:simulation-execution}}

\graphicspath{{figures/main/}}

\title{Learning a patient world model for health system simulation in intensive care}
\hypersetup{pdftitle={Learning a patient world model for health system simulation}}
\author[1,2,\ensuremath{\dagger}]{Yingtai Li}
\author[1,2,\ensuremath{\dagger}]{Ruoyan Wang}
\author[3,4,5]{Yu Li}
\author[1,2,6,7,8,\ensuremath{\ddagger}]{Shaohua Kevin Zhou}

\affil[1]{School of Biomedical Engineering, Division of Life Sciences and Medicine, University of Science and Technology of China (USTC), Hefei, Anhui, 230026, China}
\affil[2]{Medical Imaging, Robotics, Analytic Computing \& Learning (MIRACLE) Lab, YRD-RIGHT, Suzhou Institute for Advanced Research, USTC, Suzhou, Jiangsu, 215123, China}
\affil[3]{School of Population Medicine and Public Health, Chinese Academy of Medical Sciences \& Peking Union Medical College, Beijing, 100006, China}
\affil[4]{Innovation Center for Philosophy and Social Sciences in Higher Education, Innovation Center for Public Health Emergency Management, Beijing, 100006, China}
\affil[5]{State Key Laboratory of Respiratory Health and Multimorbidity, Beijing, 100006, China}
\affil[6]{Jiangsu Key Laboratory of Multimodal Digital Twin Technology, Suzhou, Jiangsu, 215123, China}
\affil[7]{State Key Laboratory of Precision and Intelligent Chemistry, USTC, Hefei, Anhui, 230026, China}
\affil[8]{Biomedical Basic Research Center (BBRC) of Jiangsu Province, Suzhou, Jiangsu, 215123, China}
\affil[\ensuremath{\dagger}]{These authors contributed equally to this work (\texttt{\{liyingtai, wry050830\}@mail.ustc.edu.cn}).}
\affil[\ensuremath{\ddagger}]{Corresponding author (\href{mailto:skevinzhou@ustc.edu.cn}{\texttt{skevinzhou@ustc.edu.cn}}).}
\date{}

\begin{document}

\maketitle
\begin{abstract}
Designing healthcare strategies requires anticipating their clinical and operational consequences before implementation, a task that often relies on health system simulation.
Such a simulation necessitates modeling how patient states evolve and generate care demands. 
However, conventional approaches typically tailor compact state transitions to a single policy question and omit history-dependent dynamics, thus requiring the redefinition of states and transitions when there is a new question that entails different clinical information. 
To bridge this gap, we propose to learn a patient world model from longitudinal electronic health records using a deep neural network  to provide a shared clinical representation of patient health trajectories that can serve many simulation questions across scenarios. 
Taking intensive care unit (ICU) operations as a working example, 
we train the model on MIMIC-IV, freeze it, and couple its time-queryable predictions to shared ICU calendars, resources, and queues via scenario-specific mappings, so that one model renders admission demand, care tasks, and discharge proposals across simulations evaluating ICU bed capacity, care dispatch, and post-ICU transfer capacity. 
Extensive experiments show that our model matches task-fitted Markov and semi-Markov comparators in aggregate occupancy and directional responses to capacity and dispatch policies. 
It also reproduces policy gains from replayed observed trajectories.
Moreover, for patients entering the same support state from different prior trajectories, the learned representation captures a 24-hour recovery difference ($-6.4$ predicted versus $-6.7$ observed percentage points) --- a contrast totally missed by compact state models --- which propagates into distinct simulated care demand, queue burden, and bed use for downstream operations.
Together, these results indicate that a learned world model of patient dynamics has the potential to replace question-specific state models, allowing new care policies to be evaluated by reconfiguring the scenario without model rebuilding. \end{abstract}

\section*{Introduction}

Designing healthcare strategies requires anticipating their clinical and operational consequences before implementation. 
Health system simulation (HSS) meets this requirement by simulating the evolution of patient health states or trajectories and linking such information to healthcare \citep{Hennessy2015POHEM,Marshall2015SIMULATE}. 
In population screening and prevention, for example, simulations link tumour onset and progression to early detection \citep{Meza2021LungScreening} or couple infection dynamics with transmission and vaccination \citep{Pitman2012DynamicTransmission}. 
In hospital operations, evaluations of triage protocols \citep{Melman2021ScarceResources}, care-pathway redesign \citep{Lebcir2017ParkinsonCarePathway} and bed allocation \citep{Han2021BedAssignment} likewise trace how individual patient trajectories generate competing demands for care over time. 
Underlying these applications that differ in their mechanisms and timescales lies a computational representation that models the evolution of patient health states in tandem with care events, such as treatments, interventions, and service delivery.

Conventional HSS approaches, however, typically rely on fit-for-purpose abstraction, in which modellers specify compact clinical states, transitions and service mechanisms tailored to a single policy question \citep{Roberts2012Conceptualizing,Marshall2015Selecting,Chen2024COPDScreening}. 
Such explicit specifications keep assumptions transparent and computation tractable, but their scope is bounded by the information retained in the chosen states. 
Although modellers can augment states with patient attributes or selected prior events, each added dependency expands the state space and requires re-estimating transition parameters from data \citep{Siebert2012StateTransition,Krijkamp2018Microsimulation}. 
Compact states specified for one question may therefore omit outcome-relevant history, and answering a new question that requires different clinical granularity --- or continuous-time alignment with a shared event clock --- often demands redesigning the state space or interpolating across a fixed time grid. This modeling rigidity calls for a shared representation of multivariable clinical histories, which ideally retains rich clinical context across questions and supplies the clinical signals and care demands each analysis requires.

Deep learning on longitudinal electronic health records (EHRs) provides a foundation for learning such a shared representation: sequence models have advanced from disease-risk prediction \citep{Li2020BEHRT,Rasmy2021MedBERT} to generative models that roll out multi-step clinical trajectories as world models \citep{Ha2018WorldModels,Kraljevic2024Foresight,Renc2024ETHOS,Shmatko2025Delphi,Waxler2025Curiosity,Mu2026EHRWorld}.
At the same time, generative agent simulations \citep{Aher2023SimulateHumans,Horton2023HomoSilicus,Park2023GenerativeAgents,Guan2025LightSociety} and clinical simulation environments \citep{Luo2026CES,Yuan2026Asclepius,Li2024AgentHospital,Qiu2025DiagGym,Kuruppu2025CriticalCareTwin,Zhong2026ICUTwin} have shown how policy consequences can be studied by coupling individual trajectories within a shared environment. 
Yet existing health trajectory generators such as ETHOS advance each patient's timeline asynchronously from one emitted event token to the next \citep{Renc2024ETHOS}. 
Using learned patient dynamics inside HSS --- where multiple patients compete for beds and staff in real time --- requires advancing trajectories to arbitrary times on a shared clock and mapping their clinical states to operational demands. 
Turning a longitudinal EHR generator into a shared simulation engine therefore requires coupling time-conditioned clinical queries with explicit demand mappings, and testing whether the retained history alters both clinical predictions and downstream operational conclusions. Such a simulation engine is yet to be built.

To this end, we propose a unified framework that learns a time-conditioned patient world model (pWM) from longitudinal EHRs to capture patient health trajectories and further couples the learned pWM into discrete-event HSS.
Much as language models represent text over a shared vocabulary, the pWM encodes irregular clinical records as sequences drawn from a structured vocabulary of physiological measurements, laboratory values, treatments and clinical events, and predicts future health states and event rates across continuous-valued time intervals so that trajectories can be queried on a shared event clock.
Because this clinical vocabulary and the trajectories composed from the pWM are far richer than the small set of mutually exclusive states typically hand-specified in Markov models, the same frozen representation can supply diverse operational signals while distinguishing patients who might appear identical under a compact state definition. 
We test this framework in an acute resource-constrained setting and connect the frozen model through scenario-specific mappings to a virtual intensive care unit (ICU), where patients compete for shared beds and staff. 
In this environment, we simulate different ICU operational scenarios
with different mappings that translate generated admission proposals into bed requests, organ-support events into staff care tasks, and live-exit proposals into downstream service requests, while configuring resource capacities, service durations, and dispatch rules for each scenario.

Using patient-level partitions of MIMIC-IV \citep{Johnson2023MIMICIV} to instantiate this virtual ICU, we first verify the clinical reliability of the frozen pWM at the individual and group levels.
Fixed-history queries accurately discriminate individual endpoints such as in-hospital mortality (AUROC 0.937) and 30-day ICU admission (0.881) with near-unit recalibrated slopes (median 0.988). Using 14-day rollouts to construct virtual cohorts, we reproduce an aggregate ICU occupancy (restricted mean stay 2.92 versus 2.80 days) alongside characterized biases in support-event rates and heterogeneity. 
We then examine two simulation questions. 
First, across separately configured admission, care-delivery, and discharge scenarios, the single frozen model matches task-fitted Markov and semi-Markov comparators in directional responses to capacity and dispatch policies, and closely tracks policy gains in an observed-trajectory replay. 
Second, for matched patients entering single-organ support from no support versus dual support, the learned representation recovers an observed 24-hour recovery difference ($-6.4$ versus $-6.7$ percentage points) missed by the compact state models and propagates that contrast into distinct simulated care demand, queues, and bed use for downstream operations. 
More broadly, our work demonstrates how a learned pWM can replace bespoke, task-specific state machines with a shared clinical engine that retains patient health trajectories across diverse simulation questions.

\begin{figure}[t]
  \centering
  \includegraphics[width=\linewidth]{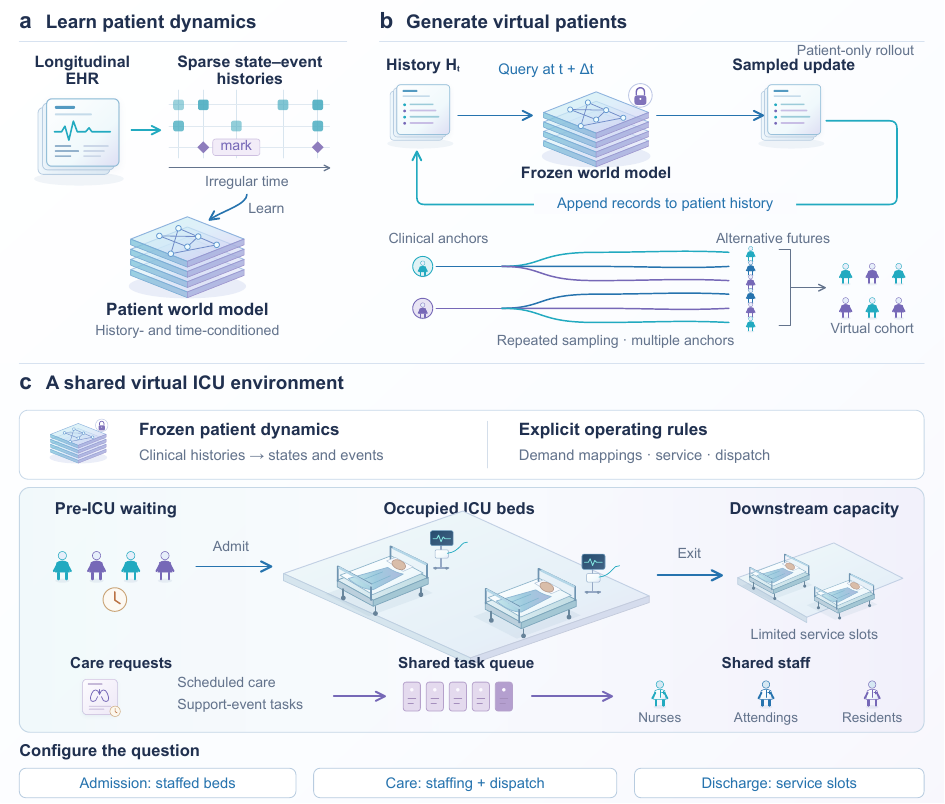}
  \captionsetup{font={footnotesize,stretch=1}}
  \caption{\textbf{A shared patient world model for different ICU simulation questions.}
  \textbf{a}, Longitudinal EHRs are encoded as sparse state--event sequences over a shared clinical vocabulary, preserving variable values, events with categorical labels and irregular observation times to train a generative patient world model conditioned on history and forecast interval $\Delta$.
  \textbf{b}, With learned parameters held fixed, querying the model at a requested future time returns state and event distributions, from which sampled updates are appended to history before the next query. Repeated rollouts from a single patient's anchor history generate alternative individual futures; rollouts across patients construct a virtual cohort.
  \textbf{c}, Scenario-specific mappings connect outputs of the frozen patient model to admission demand, staff care tasks and ICU exit within a shared discrete-event simulator linking pre-ICU waiting, beds, staff queues and downstream service slots (\cref{mfig:simulation-execution}). The schematic illustrates the full framework architecture; admission, care delivery and discharge are evaluated in separately configured scenarios. In subsequent figures, ``Neural model'' denotes this generative patient world model.}
  \label{fig:framework}
\end{figure}

\begingroup
\renewcommand{\topfraction}{0.88}
\renewcommand{\bottomfraction}{0.82}
\renewcommand{\textfraction}{0.10}
\renewcommand{\floatpagefraction}{0.90}
\setlength{\textfloatsep}{10pt plus 2pt minus 2pt}
\setlength{\intextsep}{9pt plus 2pt minus 2pt}
\setlength{\floatsep}{10pt plus 2pt minus 2pt}
\captionsetup[figure]{skip=5pt}

\section*{Results}

\subsection*{A shared model of patient dynamics drives a configurable virtual ICU}

The virtual ICU uses a shared model of patient dynamics to supply clinical trajectories across care and resource scenarios (\cref{fig:framework}).
Trained on longitudinal EHRs, the neural autoregressive model represents irregular patient histories as sparse state--event sequences over a clinical vocabulary of $64$ state variables and $25$ event types (\cref{fig:framework}a).
Dual-axis attention across clinical variables and elapsed time combines this context with continuous-valued forecast intervals, returning clinical predictions at the refresh times requested by the simulator.
During trajectory rollout, sampled states and events are appended to the patient's history before the next query: repeated sampling from a single patient's anchor history generates alternative futures for that individual, and sampling across patients at their respective anchors constructs virtual cohorts (\cref{fig:framework}b).

Scenario-specific readouts and demand mappings connect this frozen patient representation to separately configured admission, care delivery, and discharge simulations (\cref{fig:framework}c; \cref{mfig:study-mapping}).
Admission proposals request ICU beds, which remain occupied until terminal exit or --- in the discharge scenario --- until a downstream service slot completes post-ICU transfer.
Within the unit, admission and elapsed-time schedules generate routine nursing and physician tasks, while model-generated organ-support events trigger acute care tasks where configured.
Each task queues for its required staff role, occupies that capacity during service, and releases it upon completion, while a shared event clock coordinates arrivals, care requests, and service completions to record waiting, workload, utilization, and throughput.
Addressing a new operational question therefore requires changing only the demand mappings, capacity schedules, service rules or dispatch policy while keeping the learned patient representation and parameters fixed (\cref{mfig:simulation-execution}).

\FloatBarrier
\subsection*{The frozen pWM supports individual-level risk prediction across clinical endpoints}

Before using the pWM to generate virtual cohorts, we evaluate its individual-level risk predictions from fixed observed histories across six clinical endpoints, each tested on $1{,}024$ held-out records with $1{,}000$ patient-cluster bootstraps (\cref{fig:patient-validity}a,b).
All six endpoints are read from the same frozen model parameters by varying only the query anchor, target, stopping event, and horizon.
Across tasks, AUROC reaches $0.937$ for in-hospital mortality, $0.881$ for $30$-day ICU admission, $0.814$ for $30$-day ICU readmission, and $0.682$ for $30$-day hospital readmission, with precision--recall areas exceeding task prevalence across all six endpoints (\cref{fig:patient-validity}a; \cref{fig:supp-calibration}a).
Brier point estimates are lower than empirical-frequency references for five of the six endpoints, with $95\%$ confidence intervals entirely below zero for four (\cref{fig:patient-validity}b).
Incorporating additional clinical context improves both metrics: as observed ICU history increases from $0\text{ h}$ to $24\text{ h}$, ICU-mortality AUROC rises from $0.797$ to $0.874$, and the Brier difference shifts from $+0.005$ to $-0.019$.

\begin{figure}[htb]
  \centering
  \includegraphics[width=\linewidth]{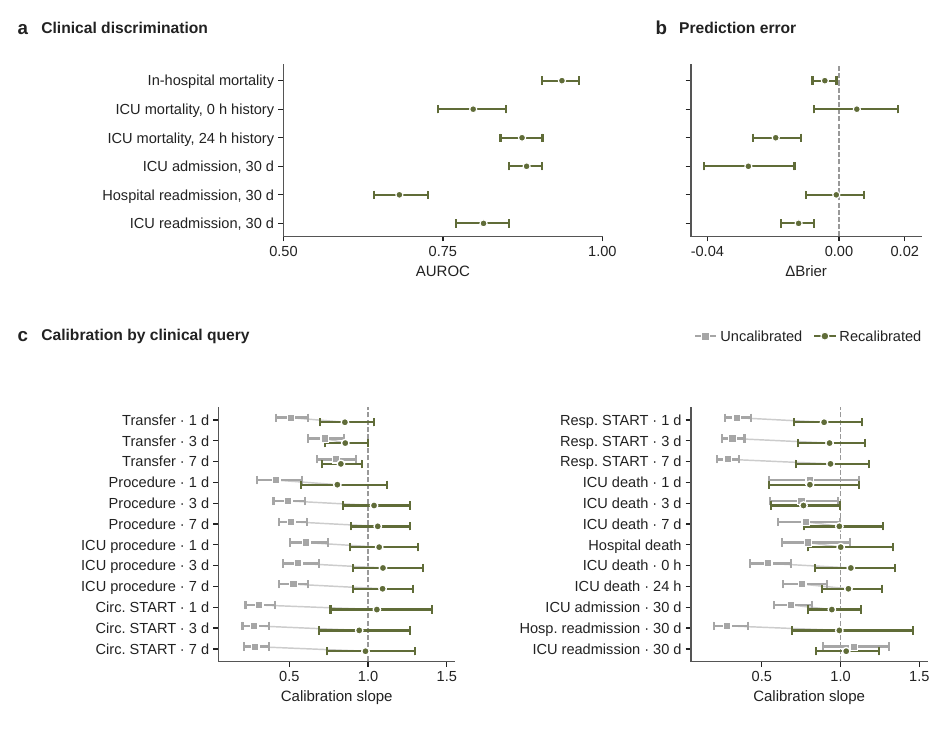}
  \captionsetup{font={footnotesize,stretch=1}}
  \caption{\textbf{Clinical queries from a frozen patient model.}
  \textbf{a}, AUROC across six clinical endpoints, each evaluated on 1,024 held-out test records from 956--981 patients. For ICU mortality, 0 and 24 h denote observed history after ICU admission.
  \textbf{b}, Brier score differences (model minus empirical-frequency reference); negative values favor the model.
  \textbf{c}, Uncalibrated and recalibrated slopes across 24 binary task--horizon queries; connecting lines pair estimates for the same query and the dashed line marks slope one. Whiskers in \textbf{a}--\textbf{c} show 95\% patient-cluster bootstrap intervals from 1,000 resamples. All queries and calibration settings share the same frozen model parameters.}
  \label{fig:patient-validity}
\end{figure}

Recalibration on validation data brings query-specific calibration slopes close to one across horizons while keeping the underlying model parameters fixed (\cref{fig:patient-validity}c).
Across $24$ binary task--horizon queries, the median recalibrated slope reaches $0.988$ (interquartile range, $0.885$--$1.051$), with $17$ of $24$ point estimates falling within $[0.9,1.1]$ and $21$ of $24$ confidence intervals including $1.0$. Recalibration also improves $72\text{-h}$ nominal $80\%$ interval coverage across $14$ state variables from $65.5\%$ to $81.4\%$ (\cref{fig:supp-calibration}b), showing that query-level recalibration can adjust probability and state readouts across tasks and horizons without retraining the underlying patient model.

\FloatBarrier
\subsection*{Model-simulated cohorts approximate population-level ICU occupancy and stay duration}

Having validated clinical predictions from observed histories, we next test whether multi-day autoregressive rollouts from the frozen model could approximate population-level ICU courses (\cref{fig:population-validity}).
We generate $16$ recalibrated $14$-day rollouts from the ICU-admission anchor of each of $670$ held-out test patients ($10{,}720$ trajectories in total) and map trajectories to five states (no support, single-organ support, dual support, live discharge, and death).
Across the $14$-day horizon, the virtual cohort closely tracks the observed ICU retention curve (\cref{fig:population-validity}b) and hourly five-state occupancy distributions (mean total variation distance $0.0388$, Jensen--Shannon divergence $0.0044$; \cref{fig:supp-trajectory}), yielding a restricted mean ICU stay of $2.923\text{ days}$ (median $1.917\text{ days}$) compared with $2.800\text{ days}$ observed (median $1.738\text{ days}$).

\begin{figure}[htb]
  \centering
  \includegraphics[width=\linewidth]{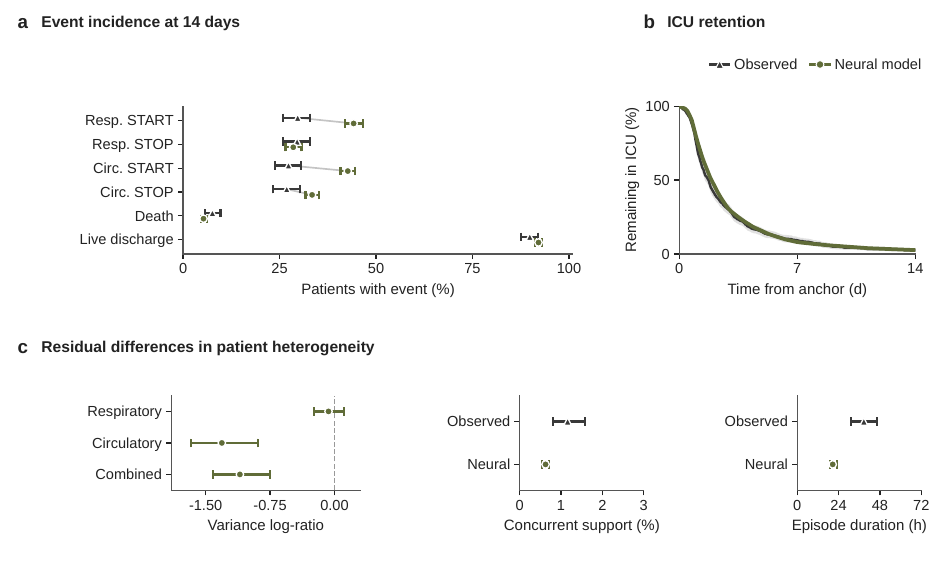}
  \captionsetup{font={footnotesize,stretch=1}}
  \caption{\textbf{Longitudinal generation of virtual ICU cohorts.}
  Sixteen recalibrated 14-day rollouts are generated from each of 670 held-out test patients at ICU admission (10,720 trajectories).
  \textbf{a}, Day-14 support-transition and terminal-event incidence; lines pair observed and model estimates.
  \textbf{b}, Fraction of patients remaining in the ICU over 14 days.
  \textbf{c}, Patient-level heterogeneity: generated-to-observed variance log-ratios for per-patient support initiations ($0$ indicates equal variance), dual-support occupancy as a fraction of follow-up patient-hours, and mean duration of respiratory episodes ending with an explicit \textsc{stop} (unadjusted for terminal censoring). Whiskers and bands show 95\% patient-bootstrap intervals from 1,000 resamples.}
  \label{fig:population-validity}
\end{figure}

Examining individual event types and patient-level variation reveals specific generative biases alongside preserved clinical structure (\cref{fig:population-validity}a,c).
Although $14$-day \textsc{start} incidence is overestimated for respiratory ($44.2\%$ versus $29.7\%$) and circulatory support ($42.7\%$ versus $27.3\%$), cumulative incidence closely tracks observations for live ICU discharge ($92.1\%$ versus $89.9\%$), ICU death ($5.3\%$ versus $7.6\%$), and respiratory ($28.6\%$ versus $29.6\%$) and circulatory \textsc{stop} events ($33.4\%$ versus $26.9\%$; \cref{fig:population-validity}a).
Concurrent dual-support occupancy is lower in generated trajectories ($0.63\%$ versus $1.16\%$ of patient-hours), and respiratory episodes ending with an explicit \textsc{stop} are shorter ($20.6\text{ h}$ versus $38.7\text{ h}$), coinciding with more episodes remaining active without an explicit \textsc{stop} until ICU exit or day $14$ ($39.95\%$ versus $10.60\%$).
Meanwhile, as measured by the generated-to-observed variance log-ratio ($0$ indicating equal variance), patient-to-patient variation in support initiations is attenuated for circulatory support ($-1.31$, 95\% CI $[-1.67,-0.89]$) but closely matches observations for respiratory support ($-0.07$, $[-0.23,0.11]$; \cref{fig:population-validity}c).

Distributional discrepancies accumulate gradually as autoregressive rollouts progress over longer event sequences (\cref{fig:supp-depth-precision}a).
At a fixed patient count, however, repeated sampling from each anchor history generates $99.8\%$ unique individual trajectories over $14\text{ days}$ and reduces Monte Carlo sampling error four-fold between $1$ and $16$ rollouts (\cref{fig:supp-depth-precision}b,c).
When benchmarked against task-fitted Markov and semi-Markov models under matched admission-state initialization in $256$ held-out stays, all three models achieve integrated mean absolute errors below $2\%$ in hourly five-state occupancy: $1.58\%$ for the neural model, $1.53\%$ for Markov and $1.24\%$ for semi-Markov (\cref{fig:supp-engine-benchmarks}a).
For support-onset timing over $7\text{ days}$, semi-Markov has the lowest Wasserstein distance for first onsets ($1.80\text{ h}$ versus $8.13\text{ h}$ for the neural model and $10.12\text{ h}$ for Markov), whereas the neural model has the lowest error across all support starts including recurrences ($3.86\text{ h}$ versus $5.41\text{ h}$ and $11.84\text{ h}$, respectively; \cref{fig:supp-engine-benchmarks}b).

\FloatBarrier
\subsection*{A single frozen pWM supports admission, care-delivery, and discharge simulations}

To test whether a shared model of patient dynamics could support multiple operational questions without state redesign, we connect the frozen pWM to admission, care delivery, and discharge simulations through scenario-specific mappings (\cref{fig:operational-reuse}; \cref{mfig:study-mapping}).
In each scenario, we compare the neural model with well-established Markov and semi-Markov state-transition models fitted to the compact support states, with all models sharing the same elapsed-time routine care schedules.
Because each model supplies its own trajectories after ICU admission and native organ-support events, quantitative estimates reflect model-specific stay durations and event rates. Across all three scenarios, the single frozen model agrees with both task-fitted baselines on the direction of operational responses to capacity and dispatch controls.

\begin{figure}[htb]
  \centering
  \includegraphics[width=\linewidth]{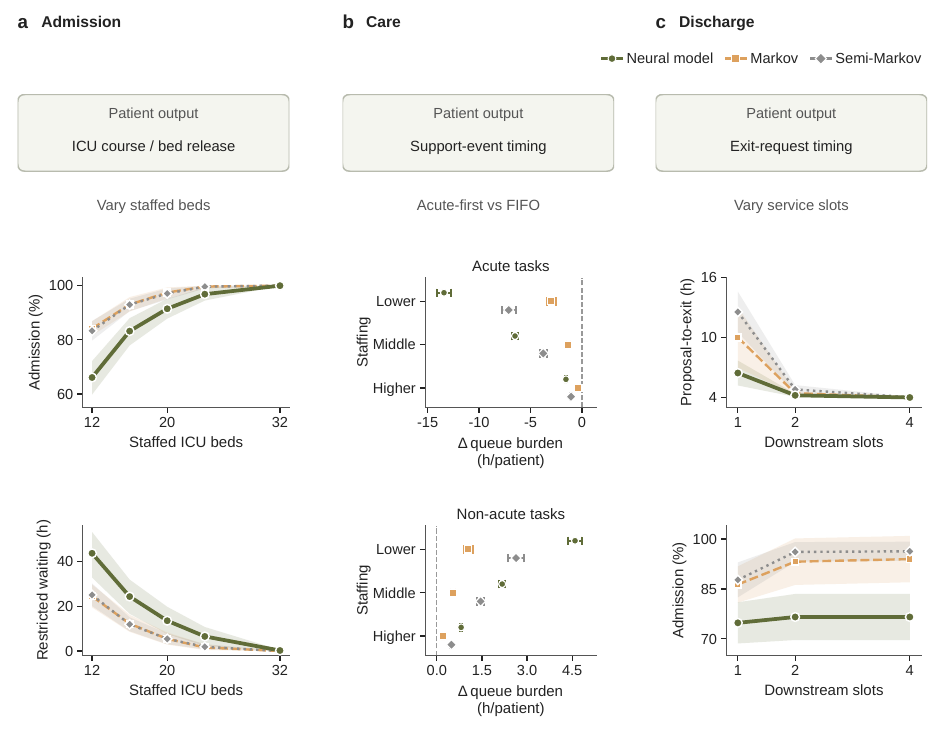}
  \captionsetup{font={footnotesize,stretch=1}}
  \caption{\textbf{Operational responses across admission, care-delivery and discharge simulations.}
  Headers above each column indicate the mapped patient-model output and the varied operational control across separately configured scenarios, comparing the frozen neural model with task-fitted Markov and semi-Markov baselines.
  \textbf{a}, ICU admission fraction and restricted pre-ICU waiting across staffed-bed capacities, sharing pre-ICU admission requests from the calibrated neural model (bands, repeat-bootstrap 95\% intervals over eight replications).
  \textbf{b}, Acute-first-minus-FIFO differences in acute and non-acute queue burden (cumulative task-waiting hours per initial patient) across staffing levels using model-specific trajectories with delivery feedback disabled ($128$ test patients, $128$ beds, $168\text{-h}$ horizon; whiskers, paired Student-$t$ 95\% intervals over $16$ replications).
  \textbf{c}, Mean time from live-exit proposal to bed release (among completed transfers) and ICU admission fraction across one to four concurrent post-ICU transfer slots with a $4\text{-h}$ transfer duration (bands, Student-$t$ 95\% intervals over eight replications).}
  \label{fig:operational-reuse}
\end{figure}

\paragraph{Admission capacity.}
Expanding staffed beds raises admission fractions and reduces pre-ICU waiting across models (\cref{fig:operational-reuse}a).
Given shared pre-ICU admission requests from the calibrated neural model, longer simulated ICU stays after admission in the neural model yield lower admission and longer restricted waiting at $12$ staffed beds: $66.1\%$ admitted and $43.63\text{ h}$ waiting versus $83.3\%$ and $25.05\text{ h}$ for semi-Markov.
Increasing capacity from $12$ to $32$ beds raises admission fractions to $100\%$ and reduces waiting to near zero in both the neural model and the baselines ($0.24\text{ h}$ versus $0.01\text{ h}$ for semi-Markov).

\paragraph{Care-delivery dispatch.}
Prioritizing acute tasks over FIFO dispatch trades longer routine waiting for a reduction in acute queue burden, with the largest gains under tight staffing (\cref{fig:operational-reuse}b).
At low staffing, acute-first dispatch reduces acute queue burden by $13.41\text{ h}$ per patient for the neural model, $7.12\text{ h}$ for semi-Markov and $2.98\text{ h}$ for Markov.
As staffing increases, these reductions converge toward zero across models ($1.56$, $1.05$ and $0.34\text{ h}$ per patient at high staffing), accompanied by small increases in non-acute waiting.

\paragraph{Downstream discharge.}
In the discharge scenario, patients proposed for live ICU exit continue to occupy their beds until completing a $4\text{-h}$ post-ICU transfer process across one, two or four concurrent transfer slots (\cref{fig:operational-reuse}c).
Expanding transfer capacity from one to four concurrent slots substantially shortens exit delays in all three models, reducing mean time from exit proposal to bed release among completed transfers from $6.45\text{ h}$ for the neural model, $10.01\text{ h}$ for Markov and $12.56\text{ h}$ for semi-Markov to $4.00$--$4.02\text{ h}$, approaching the $4\text{-h}$ service duration.
Faster bed turnover also raises ICU admission fractions across models ($74.8\%$ to $76.6\%$ for the neural model, $86.3\%$ to $93.9\%$ for Markov and $87.7\%$ to $96.3\%$ for semi-Markov), with gains plateauing beyond two slots.

\paragraph{Execution-protocol sensitivity.}
We test whether care-delivery conclusions are sensitive to the simulator's time-stepping and history-update rules without retraining the patient model (\cref{fig:supp-execution-sensitivity}b,c).
Because the model accepts continuous-valued forecast intervals, we vary the clinical refresh step across $30$, $60$ and $120\text{ min}$: although proposal-to-commit latency and simulated mortality shift with refresh frequency, mean task-start delay differs from the $60\text{-min}$ baseline by only $-0.13\text{ min}$ (95\% CI $[-0.86,0.60]$) at $30\text{ min}$ and $-0.39\text{ min}$ ($[-0.93,0.16]$) at $120\text{ min}$ (\cref{fig:supp-execution-sensitivity}b).
Enabling delivery feedback so that patient history updates upon support-task completion --- rather than replaying fixed trajectories generated without queueing --- attenuates the low-staffing acute-first reduction by $4.39\text{ h}$ per patient ($[3.79,5.00]$), while preserving both the priority benefit and its staffing gradient (\cref{fig:supp-execution-sensitivity}c).

\FloatBarrier
\subsection*{Simulated operations capture bed-and-staff bottleneck shifting and preserve observed-replay dispatch gains}

\begin{figure}[p]
  \centering
  \includegraphics[width=\linewidth]{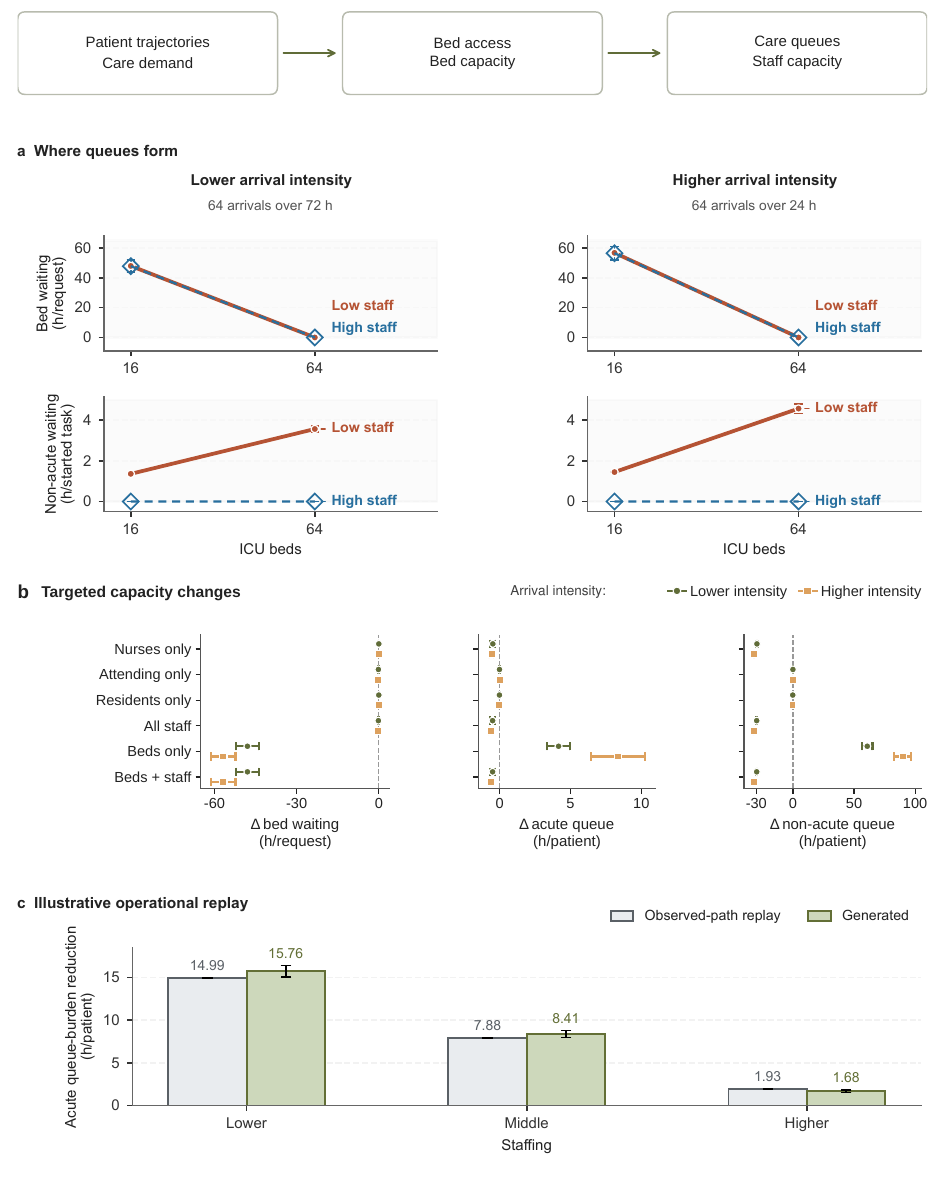}
  \captionsetup{font={footnotesize,stretch=1}}
  \caption{\textbf{Bed-and-staff capacity interactions and observed-trajectory operational replay.}
  The upper schematic illustrates how expanding bed access shifts demand into staff-constrained care queues when staffing is fixed.
  \textbf{a}, Restricted bed-request waiting (hours per request) and started non-acute task waiting (hours per started task) across bed capacities ($16$ versus $64$ beds) and staffing levels at two arrival intensities ($64$ arrivals over $72\text{ h}$ or $24\text{ h}$).
  \textbf{b}, Paired capacity-expansion effects on bed waiting and on acute and non-acute queue burden (cumulative task-waiting hours per initial patient) relative to the $16\text{-bed}$, low-staffing baseline.
  \textbf{c}, FIFO-minus-acute-first reductions in acute queue burden driven by observed versus generated trajectories for the same $128$ held-out patients ($128$ beds, $168\text{-h}$ horizon). Points in \textbf{a},\textbf{b} and bars in \textbf{c} show means; whiskers show Student-$t$ 95\% intervals across $16$ replications (conditional on the fixed cohort in \textbf{c}).}
  \label{fig:capacity-decisions}
\end{figure}

\paragraph{Bed-and-staff capacity interactions.}
In separate and joint capacity expansions, increasing bed access under fixed staffing shifts congestion from admission to care delivery, consistent with resource-bottleneck effects in hospital-flow simulations \citep{Manley2005SurgeCapacity,Hurwitz2014EDCrowding} (\cref{fig:capacity-decisions}a,b).
Across two arrival intensities ($64$ admission requests over $72\text{ h}$ or $24\text{ h}$), restricted bed waiting in a $16\text{-bed}$ unit reaches $48.0$--$56.7\text{ h}$ per request across staffing levels, and expanding to $64$ beds eliminates admission queues ($0.0\text{ h}$, reducing bed waiting by $50$--$57\text{ h}$ per request relative to the $16\text{-bed}$, low-staffing baseline).
Under low staffing, however, admitting more patients into the unit increases mean waiting per started non-acute task from $1.36$--$1.45\text{ h}$ to $3.57$--$4.57\text{ h}$ (\cref{fig:capacity-decisions}a), raising cumulative acute queue burden by $4.16$--$8.36\text{ h}$ and non-acute burden by $60.8$--$89.8\text{ h}$ per patient (\cref{fig:capacity-decisions}b).
Conversely, expanding nurses or all staff roles at $16$ beds reduces routine queue burden by $30$--$32\text{ h}$ per patient without improving bed access.
Coordinated expansion of beds and staff relieves admission and care-delivery bottlenecks simultaneously.

\paragraph{Observed-trajectory replay.}
To test whether generated trajectories preserve quantitative dispatch-policy effects relative to observed clinical courses, we replay observed and generated trajectories for the same $128$ test patients entering simultaneously at time zero under identical scheduled tasks, binary support-onset mappings and service rules (\cref{fig:capacity-decisions}c).
Prioritizing acute tasks over FIFO dispatch reduces acute queue burden by $15.76$, $8.41$ and $1.68\text{ h}$ per patient for generated trajectories at low, middle and high staffing, closely tracking the $14.99$, $7.88$ and $1.93\text{ h}$ reductions obtained when replaying observed trajectories.
Using the acute-first-minus-FIFO convention, paired generated-minus-observed differences in policy effect remain small across staffing levels: $-0.76\text{ h}$ per patient (descriptive 95\% interval $[-1.46,-0.07]$) at low staffing, $-0.53\text{ h}$ ($[-0.95,-0.12]$) at middle staffing and $+0.26\text{ h}$ ($[0.12,0.39]$) at high staffing, showing that generated trajectories reproduce both the direction and staffing gradient of the observed-trajectory policy effect under a unified demand mapping.

\FloatBarrier
\subsection*{Clinical history omitted by compact states alters recovery predictions and simulated resource use}

\begin{figure}[p]
  \centering
  \includegraphics[width=0.98\linewidth]{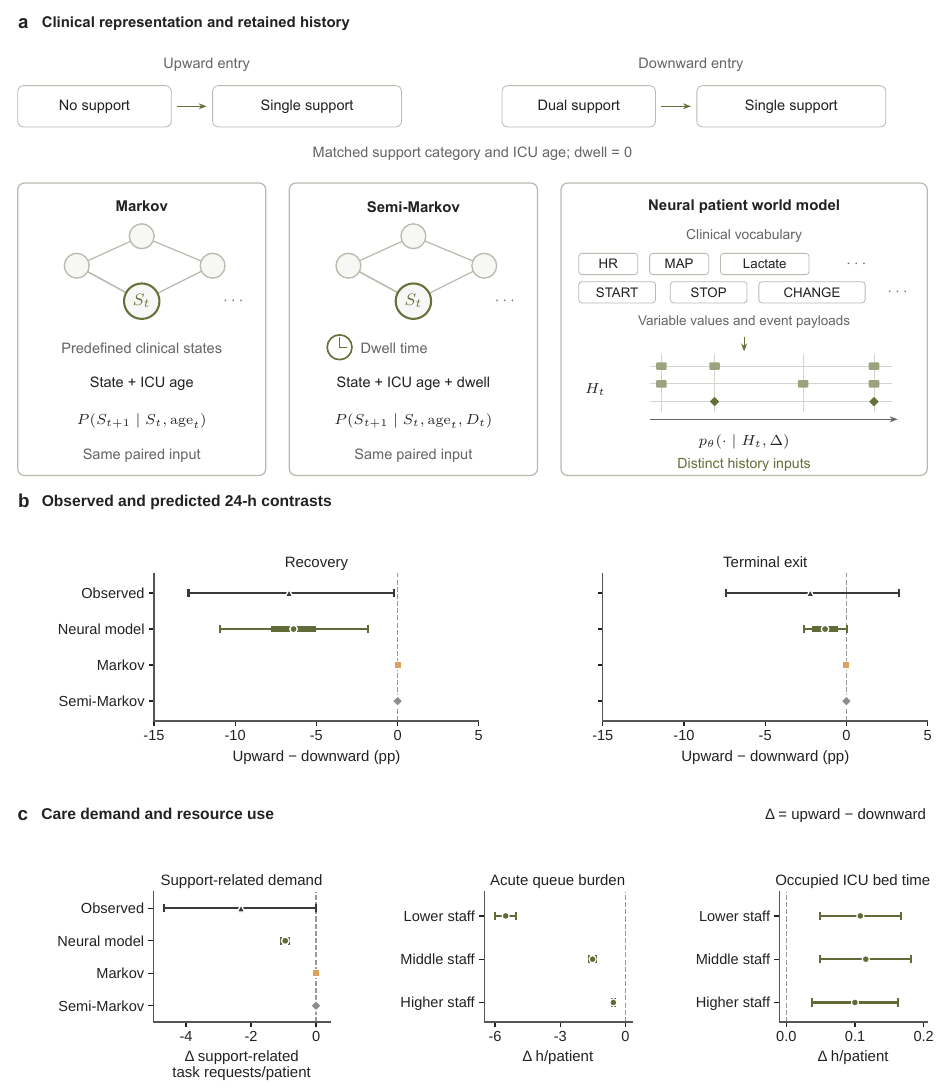}
  \captionsetup{font={footnotesize,stretch=1}}
  \caption{\textbf{Retained clinical history within a shared support state and its operational consequences.}
  \textbf{a}, Matched entry paths in $404$ held-out test pairs sharing the same single-organ support state $S_t$ and hours since ICU admission at state entry ($D_t=0$), entered either from no support (upward entry) or from dual support (downward entry). Schematics compare inputs of the Markov ($S_t$ and ICU hours), semi-Markov (adding state dwell time $D_t$, clock) and neural models (multivariable history $H_t$ and query interval $\Delta$). HR, heart rate; MAP, mean arterial pressure.
  \textbf{b}, Upward-minus-downward $24\text{-h}$ differences in recovery (reaching no-support status within $24\text{ h}$) and terminal exit (death or live ICU discharge). Thin capped whiskers show patient-sampling $95\%$ bootstrap intervals; thick segments show Monte Carlo $95\%$ intervals across $16$ paired replications.
  \textbf{c}, Upward-minus-downward differences in support-related task requests under unconstrained staffing (left), and in acute queue burden (middle) and occupied bed time (right) per patient under shared bed and staff constraints across staffing levels (observed whiskers, patient-paired $95\%$ bootstrap intervals; model whiskers, paired Student-$t$ $95\%$ intervals across $16$ replications). pp, percentage points.}
  \label{fig:history-reuse}
\end{figure}

\paragraph{Matched-history clinical contrasts.}
To test whether the shared patient representation retains outcome-relevant history omitted by the compact support states, we compare $404$ held-out patient pairs matched at state entry ($D_t=0$) on current single-organ support channel and hours since ICU admission, where one patient has just stepped up from no support (upward entry) and the other has stepped down from dual support (downward entry; \cref{fig:history-reuse}a).
Because the task-fitted Markov and semi-Markov baselines are conditioned only on current support state, dwell time in that state, and hours since ICU admission without encoding entry direction, both baselines produce identical paired trajectories and zero endpoint contrasts (\cref{fig:history-reuse}b).
By contrast, using its existing history input, the neural model closely approximates the observed $24\text{-h}$ recovery difference between upward- and downward-entry patients: $-6.40$ percentage points (pp; 95\% CI $[-10.94,-1.84]$) in the model versus $-6.68\text{ pp}$ ($[-12.87,-0.25]$) in observations, alongside terminal-exit differences of $-1.31\text{ pp}$ ($[-2.63,+0.05]$) versus $-2.23\text{ pp}$ ($[-7.43,+3.22]$).

\paragraph{Propagation to care demand and resource use.}
These retained clinical differences propagate into simulated care demand and resource utilization (\cref{fig:history-reuse}c).
Under unconstrained staffing, the neural model reproduces the negative direction of the observed upward-minus-downward difference in support-related task requests ($-0.95$ requests per patient, 95\% CI $[-1.07,-0.83]$, versus $-2.31$ observed, $[-4.66,-0.01]$).
Although the model underestimates total mapped demand and shows an opposite-sign contrast for \textsc{start} triggers, its negative overall contrast is driven by fewer \textsc{stop}- and \textsc{change}-related requests in the upward-entry group, matching the dominant source of the observed contrast ($-1.04$ versus $-2.06$ deduplicated \textsc{stop}-or-\textsc{change} requests per patient; \cref{fig:supp-history-task-decomposition}).
When coupled with shared beds and staff (\cref{fig:history-reuse}c), this lower acute task demand reduces support-related queue burden for upward-entry relative to downward-entry patients by $5.51\text{ h}$ per patient ($[-5.98,-5.04]$) at low staffing and $0.55\text{ h}$ ($[-0.62,-0.48]$) at high staffing, while a slightly longer ICU retention increases occupied bed time by $0.10$ to $0.12\text{ h}$ per patient across staffing levels --- demonstrating that clinical history omitted by compact state definitions can alter downstream operational estimates.

\FloatBarrier
\endgroup

\section*{Discussion}

This study demonstrates that a single pWM can be coupled to scenario-specific demand mappings and resource rules to support diverse HSS questions.
Across ICU admission, care delivery, and discharge simulations in MIMIC-IV, the frozen model agrees in direction with task-fitted Markov and semi-Markov baselines.
Its generated trajectories also preserve the dispatch-policy gains obtained when replaying observed patient courses.
Moreover, when patients enter the same support state from different prior trajectories, the shared representation captures an observed recovery difference that the compact states could not distinguish.
Propagating these trajectories through the simulator translates that clinical contrast into distinct care demand, queue burden, and bed use.
Decoupling learned patient dynamics from explicit system mechanics therefore allows multiple operational questions to share a single clinical engine without task-specific state redesign (\cref{mfig:study-mapping}).

Our findings also show why evaluating a pWM for HSS goes beyond single-endpoint risk prediction.
High discrimination and calibration on fixed-history queries (\cref{fig:patient-validity}) do not prevent specific event-rate biases during multi-day autoregressive rollouts (\cref{fig:population-validity}a,c).
Yet, across admission, care delivery, and discharge simulations, the frozen model still matches the directional responses of task-fitted Markov and semi-Markov baselines (\cref{fig:operational-reuse}), captures bed-and-staff bottleneck shifting, and preserves the acute-priority dispatch gains of replayed observed trajectories (\cref{fig:capacity-decisions}).
Notably, generated cohorts closely track observed ICU retention, support-state occupancy, and recurrent acute-onset timing (\cref{fig:population-validity}b; \cref{fig:supp-engine-benchmarks}), suggesting that fidelity on these aggregate and timing dimensions can support consistent operational comparisons even when marginal event counts are imperfect.
Evaluating a pWM for HSS should therefore pair cohort occupancy checks with observed-trajectory operational replay, testing directly whether generated demand preserves policy rankings under target resource rules.

Retaining multivariable clinical history also avoids manual state expansion when patient courses depend on prior trajectory rather than current state alone.
In standard state-transition modelling, accounting for an omitted historical factor requires explicitly augmenting the state space --- such as adding dwell time in a semi-Markov model or adding entry direction (\cref{fig:history-reuse}a,b) --- and re-estimating transitions \citep{Siebert2012StateTransition,Krijkamp2018Microsimulation}.
Worse, modellers must anticipate which historical variables matter before constructing the state space.
A pWM instead represents such dependencies through shared parameters over longitudinal histories, reducing the need to enumerate history-specific states.
In our matched-history comparison, the frozen model recovers the lower $24\text{-h}$ recovery rate and fewer support-related task requests of upward-entry patients without an entry-direction covariate or parameter refitting, translating this retained clinical history into distinct downstream queue burdens and bed use (\cref{fig:history-reuse}b,c).

This design bridges classical HSS and emerging longitudinal patient models.
Micro-simulation, discrete-event simulation, and agent-based models have long composed individual trajectories within shared resource environments \citep{Krijkamp2018Microsimulation,Karnon2012DES,Bonabeau2002ABM,Han2021BedAssignment}, including reusable disease platforms such as POHEM and COPD-SIM \citep{Hennessy2015POHEM,Chen2024COPDScreening}, but typically rely on hand-specified state transitions.
Conversely, longitudinal modelling of clinical records and medical claims has progressed from risk prediction and disease-signature discovery \citep{Li2020BEHRT,Rasmy2021MedBERT,Urbut2026ALADYNOULLI,Ma2026ReClaim,Hager2026RisQ,Guo2026HealthFound} to generative patient world models that roll out multi-step clinical trajectories across discrete event tokens or continuous time \citep{Kraljevic2024Foresight,Renc2024ETHOS,Shmatko2025Delphi,Waxler2025Curiosity,Mu2026EHRWorld,Lutsker2026HealthFormer,Susetzky2026NOAH,Wang2026HealthFlux}.
Yet these longitudinal generators evaluate individual patient trajectories in isolation.
Advancing multiple patients on a shared resource clock requires querying clinical states at arbitrary simulation times --- rather than only at asynchronously emitted event tokens \citep{Renc2024ETHOS} --- and translating those clinical updates into competing requests for beds, staff, and downstream services.

Coupling learned patient trajectories through shared resource queues also distinguishes our environment from recent LLM-based simulations and clinical agent testbeds.
Generative social simulations \citep{Aher2023SimulateHumans,Argyle2023OutOfOneMany,Horton2023HomoSilicus,Park2023GenerativeAgents,Guan2025LightSociety} and virtual clinical environments such as Agent Hospital, DiagGym, CES and Asclepius \citep{Li2024AgentHospital,Qiu2025DiagGym,Luo2026CES,Yuan2026Asclepius} use language models to simulate human behaviour or clinician--patient dialogues for agent evaluation.
In our environment, by contrast, patients follow stochastic clinical trajectories learned from longitudinal EHRs, and their interactions arise through competition for scarce hospital capacity.
In this respect, the virtual ICU complements health-system digital twins that link patient or population models to operational policy environments \citep{Liew2026CognitiveTwin,Kuruppu2025CriticalCareTwin,Zhong2026ICUTwin}, providing a learned clinical engine that could be paired with live operational data streams and local validation in a deployed digital twin \citep{Trauer2020DigitalTwin,Katsoulakis2024DigitalTwins}.
With explicit observation and action interfaces, this environment could also support future evaluations of operational agents that prioritize care or allocate resources.

Two primary limitations define the scope of our findings.
First, because MIMIC-IV records clinical trajectories rather than empirical queue logs or staff schedules, our simulator specifies arrival streams, service durations, and capacities from published ranges and scenario designs.
While predicting absolute waiting time or deploying the model across hospitals would require calibrating demand mappings to measured workflow data and external validation \citep{Pollard2018EICU,Collins2024TRIPODAI}, applying identical arrival, service and staffing rules across all compared arms allows our baseline, replay and matched-history comparisons to isolate differences driven by the underlying patient trajectories.
Second, like Markov and semi-Markov models fitted to observational records, the pWM learns conditional associations under historical care practices instead of causal effects \citep{Petersen2012Positivity,DAmour2021Overlap,Yu2020MOPO}; rather than estimating how delayed care alters patient survival, our simulations quantify how capacity and dispatch policies shape admission access, queue waiting, and resource utilization under observed care patterns.

Despite these constraints, our findings show that a single patient world model learned from longitudinal EHRs can serve as a shared clinical engine across diverse operational simulations.
Within the virtual ICU, coupling this frozen representation to explicit demand mappings and resource queues supports admission, care delivery, and discharge evaluations while preserving history-dependent recovery and workload differences omitted by compact states.
Because clinical dynamics are decoupled from scenario mechanics, evaluating new triage rules, care pathways, or capacity schedules requires reconfiguring only the external mappings and resource rules rather than redesigning the patient state space.
More broadly, replacing bespoke state-transition abstractions with a shared, history-preserving pWM offers a generalizable foundation for anticipating the clinical and operational consequences of health-system interventions before implementation.

\section*{Methods}

\subsection*{Study design and data}

We train a patient world model on longitudinal electronic health records (EHRs) from MIMIC-IV v3.1 \citep{Johnson2023MIMICIV} and test whether it could supply clinical dynamics across diverse HSS scenarios.
We first evaluate individual-level clinical predictions and generate cohort trajectories.
Next, we couple the frozen model to discrete-event simulations of ICU admission, care delivery, and post-ICU discharge alongside Markov and semi-Markov comparators.
Finally, we conduct a matched-history analysis to test whether the model captures history-dependent trajectory differences missed by compact state models and whether these differences translate into distinct simulated care demand and resource use.

MIMIC-IV contains de-identified hospital and ICU records from Beth Israel Deaconess Medical Center in Boston, Massachusetts, collected between 2008 and 2022 (364,627 patients and 94,444 ICU stays processed).
The institutional review board of Beth Israel Deaconess Medical Center approves the creation and sharing of the database and waives the requirement for informed consent.

We partition patients deterministically, assigning 80\% of patients to model training and keeping all records from each patient within a single partition.
Among the remaining 20\% of patients, a validation subset is used to monitor training and adjust readout recalibration parameters.
All other patients form the held-out test set, which supplies the evaluation cohorts for clinical prediction, population trajectories, comparator benchmarks, and operational simulations.
Each analysis uses a task-specific clinical anchor and follow-up horizon.

\subsection*{State--event representation}

To represent irregular longitudinal EHRs, we encode each patient's history over a clinical vocabulary of 64 state variables and 25 event types.
Records are arranged on a sparse two-dimensional grid whose columns represent timestamps (rounded to seconds) and slots identify clinical state variables or event types (\cref{mfig:patient-world-model}a).
Continuous state measurements are discretized into ten quantile bins based on observed training values, leaving missing measurements unrecorded.
Each event entry records an event occurrence and its associated categorical labels, such as medication or diagnosis codes.
Concurrent labels of the same event type within a stay are grouped into a single entry.

To limit the size of the event-label vocabularies, we map medications to Anatomical Therapeutic Chemical (ATC) codes and diagnoses and procedures to coarser International Classification of Diseases (ICD) categories where available.
Within each event type, labels covering 99\% of occurrences are retained, and the remaining labels are assigned to a residual ``other'' category.
Observed state measurements and events form sparse context entries, whereas future time columns $\tau = t + \Delta$ contain query entries specifying the clinical targets to be predicted.

\subsection*{Time-conditioned patient world model}

The pWM maps a patient's sparse state--event history and a requested forecast interval to future clinical distributions (\cref{mfig:patient-world-model}a).
For patient $i$, let $H_{i,t}$ denote the observed records and model-generated updates committed to history through time $t$.
At forecast interval $\Delta>0$, the model parameterizes
\begin{equation}
  p_{\theta}\!\left(x_{i,t+\Delta},e_{i,t+\Delta},m_{i,t+\Delta}
  \mid H_{i,t},\Delta\right),
  \label{eq:patient-kernel}
\end{equation}
where $x$ denotes discretized clinical states, $e$ event occurrences or intensities, and $m$ conditional event labels.

\begin{methodfigure}[t]
  \centering
  \includegraphics[width=\linewidth,keepaspectratio]{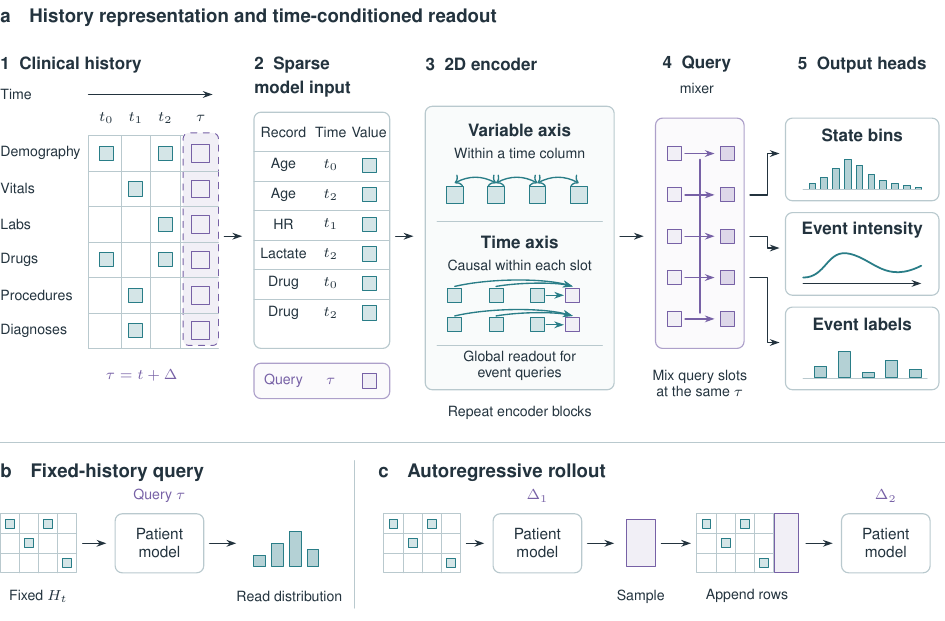}
  \captionsetup{font={footnotesize,stretch=1}}
  \caption{\textbf{Patient world model architecture, fixed-history queries and autoregressive rollout.}
  \textbf{a}, Irregular clinical records and future query columns ($\tau=t+\Delta$) are packed into a sparse model input. Two-dimensional encoder blocks alternate variable-axis attention within each time column and causal time-axis attention along each slot (with global history readout for event queries). A cross-slot mixer combines query representations at $\tau$ before output heads decode state-bin probabilities, event intensities and conditional event-label distributions.
  \textbf{b}, A fixed-history query predicts distributions at $\tau$ from observed history $H_t$ without modifying the context.
  \textbf{c}, Autoregressive rollout samples clinical records at $\Delta_1$, appends them to history and conditions the next query at $\Delta_2$ on the updated history. HR, heart rate.}
  \label{mfig:patient-world-model}
\end{methodfigure}

To embed each non-empty grid entry, the model sums learned representations for its clinical variable or event type, its discretized value,  and whether the entry is a state measurement, an event occurrence or a prediction query. Event entries also add the mean-pooled embedding of their categorical labels and a coarse event-origin embedding (grouping the 25 event types into clinical observations, care processes, clinical outcomes, administrative records, timeline boundary markers, and other events).
Column timestamps are encoded using 32 log-scale Fourier features projected to the hidden dimension.
The Transformer encoder comprises six layers with 12 attention heads, a hidden dimension of 768, a feed-forward dimension of 3,072 and dropout of 0.1.

Within each time column, variable-axis attention mixes context entries to represent contemporaneous relationships among clinical variables.
Along each slot's timeline, time-axis attention reads preceding context entries and same-column observations when present, while event queries additionally attend to all causal context entries across slots.
Within each query column, a cross-slot mixer then combines the queried slot representations at that future time before final layer normalization.

Separate output heads decode ten-bin state distributions, event intensities, and conditional event-label distributions (\cref{mfig:patient-world-model}a).
On event-query entries, a shared scalar projection directly predicts type-specific intensity logits, while type-specific categorical heads predict event labels.
Because queries accept any positive interval $\Delta$, the model supports both fixed-history risk readouts and autoregressive trajectory rollouts (\cref{mfig:patient-world-model}b,c).
A fixed-history query returns state, event, and event-label distributions at the requested future time without modifying the observed context.
During autoregressive rollout, sampled events, event labels, and, where enabled, states are appended to $H_{i,t}$ before the next query, so subsequent predictions are conditioned on the updated history.

\subsection*{Training objectives and optimization}

To supervise both step-by-step clinical evolution and competing clinical outcomes, we train the model by conditioning on observed patient histories (teacher forcing).
State targets use ten-bin cross-entropy, and event-label loss is categorical cross-entropy averaged over observed event labels.
For event type $e$, the predicted intensity logit $z_e(\tau)$ defines a positive rate:
\[
\lambda_e(\tau)=\operatorname{softplus}\!\left(z_e(\tau)\right)+\epsilon .
\]
Over an interval of width $\Delta t$ in days, observed event counts $c_e$ contribute the event loss
\[
\mathcal L_{\mathrm{event}}
=
\frac{
\Delta t \sum_{e\in\mathcal E_{\mathrm{sup}}}\lambda_e(\tau)
-
\sum_{e\in\mathcal E_{\mathrm{sup}}} c_e \log \lambda_e(\tau)
}{
\max\!\left(1,\sum_{e\in\mathcal E_{\mathrm{sup}}} c_e\right)
},
\]
where $\mathcal E_{\mathrm{sup}}$ is the set of supervised event types, giving the next-time prediction loss $\mathcal L_{\mathrm{next}}=\mathcal L_{\mathrm{state}}+\mathcal L_{\mathrm{event}}+\mathcal L_{\mathrm{label}}$.

An auxiliary target-before-stop loss supervises whether a clinical target occurs before a competing stopping event across phase-specific anchor, target, and stop definitions.
On a query grid with interval widths $\Delta t_k$, we treat the aggregated target and stop intensities $\lambda_{T,k}$ and $\lambda_{S,k}$ as constant within each interval.
With the definition of  $\eta_k=\lambda_{T,k}+\lambda_{S,k}$, the survival and interval event probabilities are given as
\begin{equation}
  \begin{aligned}
    S_k &= S_{k-1}\exp(-\eta_k\Delta t_k), \qquad S_0=1, \\
    p_{m,k} &= (S_{k-1}-S_k)\frac{\lambda_{m,k}}{\eta_k},
    \qquad m\in\{T,S\},
  \end{aligned}
  \label{eq:competing-hazards}
\end{equation}
where $S_k$ is the probability that neither event has occurred through interval $k$, and $p_{m,k}$ is the probability that event $m$ occurs first in that interval.
The endpoint-time likelihood assigns $p_{T,K}$, $p_{S,K}$ or $S_K$ to target-first, stop-first, or right-censored labels, respectively, and we minimize its negative log-likelihood $\mathcal L_{\mathrm{endpoint}}$.
We partition multi-interval queries $[t, t+H]$ at log-spaced boundaries $\tau_k = t + H\bigl[(1+s)^{k/K}-1\bigr]/s$ ($k=0,\dots,K$; $s=31$), using $K=4$ intervals for $\mathcal L_{\mathrm{endpoint}}$ during training and $K=21$ intervals when sampling next-event times over 14 days during trajectory generation.
For fixed-history clinical predictions, cumulative target and stop probabilities through horizon $H$ are obtained by summing $P_T=\sum_{k=1}^K p_{T,k}$ and $P_S=\sum_{k=1}^K p_{S,k}$, with survival probability $S_K$.
The combined training objective is
\begin{equation}
  \mathcal L =
  \mathcal L_{\mathrm{next}}+
  \alpha_{\mathrm{endpoint}}\mathcal L_{\mathrm{endpoint}},
  \label{eq:training-objective}
\end{equation}
with $\alpha_{\mathrm{endpoint}}=0.10$.

Because routine measurements vastly outnumber clinical transitions and acute interventions in raw EHR timelines, training windows are drawn using stratified sampling to upweight admission and discharge landmarks and organ-support initiations alongside random cutoffs.
We train the model using AdamW ($\beta_1=0.9$, $\beta_2=0.95$, weight decay 0.1), bfloat16 precision, gradient-norm clipping at 0.5, and a batch size of 64.
The learning rate warms up linearly over 2,000 updates to $2\times10^{-5}$ and decays following a cosine schedule to $5\times10^{-6}$ at 60,000 updates.
Each example retains a history window of at most 1,536 time columns and 3,072 grid entries.

All evaluations use the frozen 60,000-update checkpoint.
Holding the model weights fixed, we fit lightweight recalibration parameters on the validation split: logistic intercept and slope parameters for binary endpoint probabilities, variable-specific class biases and temperature scalars for ten-bin state distributions (interpolated on a logarithmic horizon scale during trajectory generation), and piecewise phase-specific rate multipliers (supplemented by support-duration multipliers) for event intensities.
These calibration parameters are held fixed across all compared conditions within each experiment.

\subsection*{Clinical prediction evaluation}

To assess individual-level discrimination and calibration from fixed histories, we evaluate six clinical prediction tasks, each using 1,024 test records: hospital mortality, ICU mortality at admission, ICU mortality at 24 h, 30-day ICU admission, 30-day hospital readmission, and 30-day ICU readmission.
These cohorts contain 981, 959, 956, 974, 961 and 965 distinct patients, respectively.
Each mortality task predicts whether death occurs before live discharge from the corresponding hospital or ICU stay, using the history observed up to hospital admission, ICU admission, or 24 h after ICU admission (among patients remaining in the ICU at 24 h).

The three 30-day tasks predict the occurrence of events within 30 days after the landmark and before competing stopping events.
Specifically, 30-day ICU admission is predicted from hospital admission (stopped by hospital discharge or death), 30-day hospital readmission from live hospital discharge (stopped by death) and 30-day ICU readmission from the first live ICU discharge within an adult hospitalization (requiring readmission to the ICU during the same hospitalization before hospital discharge or death).

Discrimination and probability accuracy are measured using the area under the receiver operating characteristic curve (AUROC), the area under the precision--recall curve (AUPRC; \cref{fig:supp-calibration}a), and the Brier score, using validation-set event prevalence as the Brier reference.
To evaluate calibration across horizons, we estimate logistic calibration slopes across 24 binary task--horizon combinations \citep{VanCalster2019Calibration} and empirical coverage of nominal 80\% prediction intervals for 14 state variables at 1, 6, 24 and 72 h using the randomized probability integral transform (PIT; proportion of PIT values between 0.1 and 0.9; \cref{fig:supp-calibration}b).

\subsection*{Population trajectory generation and evaluation}

To test whether autoregressive rollouts reproduce cohort-level ICU dynamics over 14 days, we generate trajectories from ICU admission for 670 test patients.
At each step, the model samples the next clinical event from the predicted intensities along with the 64 state variables at that event time.
Organ-support rules permit support initiation (\textsc{start}) only when the corresponding support is inactive, and parameter adjustment (\textsc{change}) or discontinuation (\textsc{stop}) only when it is active.
Each rollout terminates at death, live ICU discharge, day 14 or 256 generated events.
We generate 16 raw and 16 recalibrated rollouts per patient (10,720 trajectories per variant), averaging trajectories within each patient before weighting patients equally.

Observed and generated trajectories are projected onto five mutually exclusive states: no support, single-organ support (respiratory or circulatory), dual respiratory and circulatory support, live ICU discharge, and death, with both terminal states absorbing (redundant \textsc{start} or \textsc{stop} events do not alter support status).
Across the 14-day horizon, we compare organ-support event rates, hourly state occupancy, terminal outcomes, and restricted ICU length of stay.
Event incidence measures the proportion of patients with at least one event, whereas event burden counts total occurrences per patient including recurrences.
Restricted ICU length of stay measures time spent in the ICU up to death, live discharge, or day-14 censoring.
To assess patient-level heterogeneity, we evaluate between-patient variance in cumulative \textsc{start} counts, the fraction of follow-up spent on dual support, and respiratory episode durations stratified by whether the episode ends in an explicit \textsc{stop} or via death, discharge, or day-14 censoring.

\subsection*{Markov and semi-Markov comparators}

To benchmark the neural model against compact state representations, we fit Markov and semi-Markov models to hourly training transitions across the same five-state support projection.
Markov transition probabilities are conditioned on current support state and hours since ICU admission ($A_t$, discretized into $0$--$24$, $24$--$72$, $72$--$168$ and $\ge 168\text{ h}$) and are estimated from empirical transition counts.
Semi-Markov transition probabilities are additionally conditioned on dwell time in the current state ($D_t$, discretized into $0$--$1$, $1$--$3$, $3$--$6$, $6$--$12$, $12$--$24$, $24$--$48$ and $\ge 48\text{ h}$), regularized by adding 50 pseudo-counts distributed according to the corresponding Markov transition distribution.

We compare the three models over 14 days on 256 test patients initialized from their observed ICU admission states (\cref{fig:supp-engine-benchmarks}).
We average neural estimates of four rollouts per patient and compute Markov and semi-Markov trajectories as exact hourly analytic expectations initialized at $D_t=0$.

To evaluate the timing of organ-support onset over seven days, we restrict the 256-patient cohort to the 244 patients who receive neither respiratory nor circulatory support at ICU admission (\cref{fig:supp-engine-benchmarks}b).
We compare observed and model-predicted timing distributions using 1-Wasserstein distances (in hours) for both first support onset and all support \textsc{start} events including recurrences.

\subsection*{Care-demand and resource specification}

\begin{methodfigure}[!ht]
  \centering
  \includegraphics[width=\linewidth,keepaspectratio]{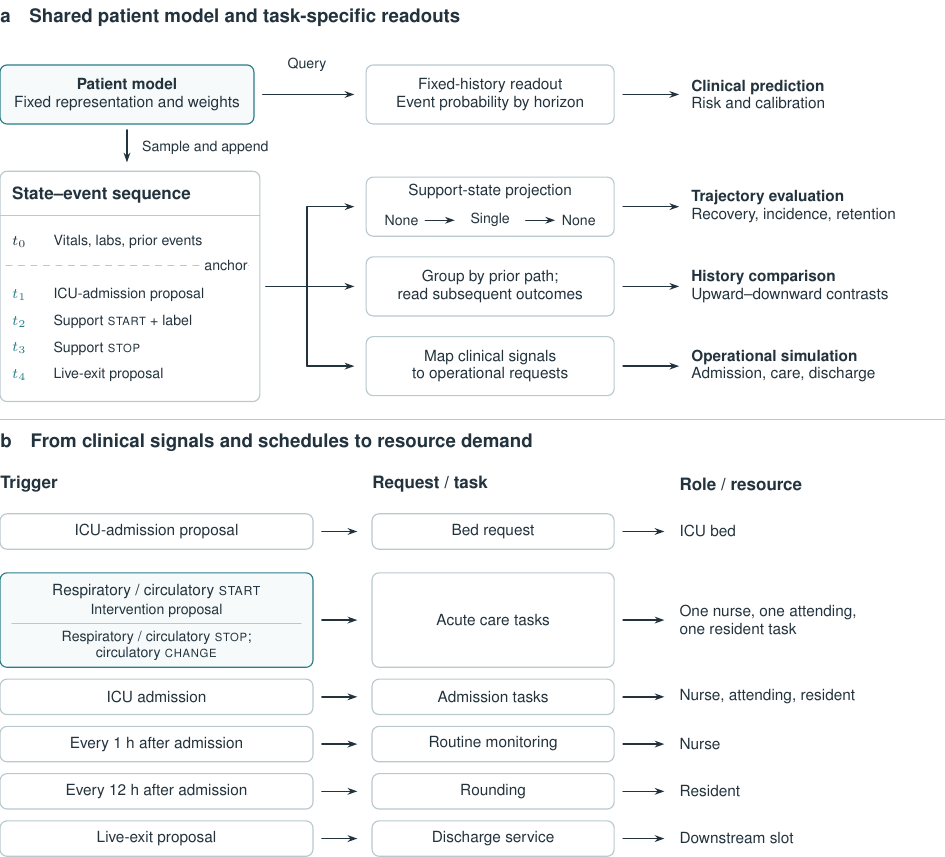}
  \captionsetup{font={footnotesize,stretch=1}}
  \caption{\textbf{Shared patient-model readouts and resource-demand mappings.}
  \textbf{a}, With representation and weights held fixed, querying the patient model directly provides fixed-history risk readouts for clinical prediction (top), whereas sampling and appending records generates state--event sequences ($t_1$--$t_4$) that support trajectory evaluation (via five-state support projection), matched-history comparisons and operational simulations (bottom).
  \textbf{b}, Translation of clinical triggers and elapsed-time schedules into hospital resource requests. ICU-admission proposals request an ICU bed; organ-support events (\textsc{start} proposals, \textsc{stop} signals and circulatory \textsc{change} signals) trigger acute care tasks requiring one nurse, one attending physician and one resident; ICU admission and elapsed-time schedules (every 1 h and 12 h) generate routine admission, monitoring and rounding tasks; and live-exit proposals request a downstream discharge slot.}
  \label{mfig:study-mapping}
\end{methodfigure}

To reuse the frozen patient model across operational scenarios, we utilize scenario-specific mappings to convert selected model outputs into hospital service requests without altering the learned representation or parameters (\cref{mfig:study-mapping}). Here, 
admission proposals, organ-support events, and live-exit proposals serve as clinical inputs, and the mappings define their operational requests.

\paragraph{Care-demand mapping.}
All patient models share the same admission and routine-care schedules.
An ICU admission generates one task each for a nurse, an attending physician, and a resident, followed by hourly nurse monitoring and 12-h resident rounding.

Organ-support signals generated by the model trigger acute care tasks, with each trigger requesting one nurse, one attending physician, and one resident.
In the Markov and semi-Markov comparators, acute tasks are triggered whenever organ support escalates from no support to single-organ or dual support, or from single-organ to dual support.
In the neural model, respiratory and circulatory \textsc{start} events are intervention proposals whose clinical delivery requires completion of the attending physician's acute task, whereas circulatory \textsc{change} and \textsc{stop} events and respiratory \textsc{stop} events are recorded directly as clinical signals.

\paragraph{Service times.}
Each care task requires one staff member of its designated role for a service duration sampled from a triangular distribution (\cref{fig:supp-execution-sensitivity}a).
Minimum and maximum durations are taken from published ICU workflow ranges \citep{Zhong2026ICUTwin}, with their midpoint as the mode, and are held fixed across patient models.

\paragraph{Resource pools.}
ICU beds and the three staff roles form separate integer-capacity pools.
An admitted patient occupies a bed until death or post-ICU exit, including any waiting and service time for downstream transfer.
A care task occupies one staff member from the start of service until completion or cancellation.
Patients compete through these shared bed and staff pools while clinical forecasts are conditioned on each patient's own history.

\subsection*{Discrete-event simulation and delivery feedback}

\begin{methodfigure}[!ht]
  \centering
  \includegraphics[width=\linewidth,keepaspectratio]{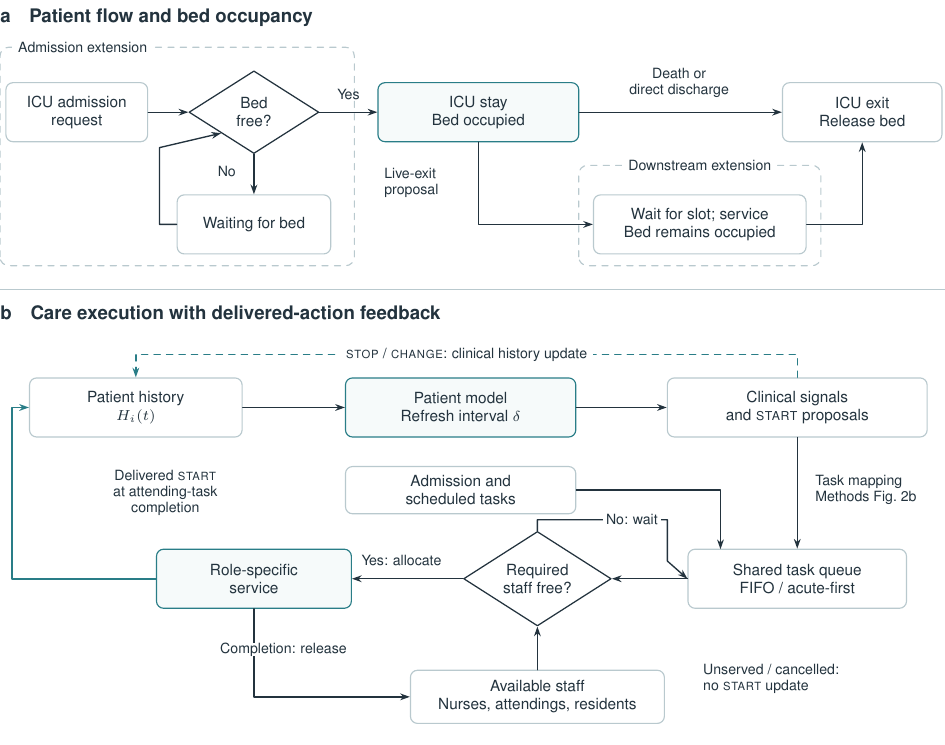}
  \captionsetup{font={footnotesize,stretch=1}}
  \caption{\textbf{Discrete-event simulation of ICU bed flow, care delivery and delivery feedback.}
  \textbf{a}, Patient flow and bed occupancy. Admitted patients occupy an ICU bed until death or discharge, with dashed boxes indicating optional pre-admission bed queueing (left) and post-ICU transfer waiting and service (right), during which the ICU bed remains occupied.
  \textbf{b}, Care execution and delivery feedback in the neural model. Scheduled care and model-generated signals enter a shared task queue (ordered by first-in, first-out (FIFO) or acute-first dispatch) and allocate available nurses, attending physicians and residents until service completion or cancellation. Whereas \textsc{stop} and \textsc{change} signals update the patient history $H_{i,t}$ directly at each clinical update (dashed teal arrow), a proposed \textsc{start} intervention is appended to $H_{i,t}$ only when its corresponding attending-physician task completes (solid teal arrow); unserved or cancelled proposals produce no history update.}
  \label{mfig:simulation-execution}
\end{methodfigure}

To couple patient trajectories to shared hospital resources, we embed the patient models within a discrete-event simulator (\cref{mfig:simulation-execution}).
The event calendar processes patient arrivals and service completions in continuous time, while each patient's clinical trajectory is updated at fixed hourly intervals ($\delta=1\text{ h}$ unless otherwise specified).
At each scheduled update, the model queries the time elapsed since the latest record in that patient's history.
Terminal outcomes are evaluated first, and a non-terminal event is sampled only if neither death nor live discharge occurs.
When events coincide in time, the simulator processes completed tasks and delivered interventions before updating clinical trajectories, generating scheduled care requests, and handling new arrivals.
Available beds and staff are then assigned to waiting requests.

Each patient enters the simulation with the clinical history observed up to entry.
When delivery feedback is enabled, a sampled \textsc{start} proposal is withheld from $H_{i,t}$ until the corresponding attending-physician task completes.
The delivered intervention is then appended to $H_{i,t}$ at its completion time, so subsequent clinical updates are conditioned on the actual time of care delivery.
Death or ICU discharge cancels all remaining tasks and pending proposals, and unserved proposals, routine task completions, and queue lengths do not modify $H_{i,t}$.
When delivery feedback is disabled, we first generate clinical trajectories under staff capacities sufficient to eliminate queueing while preserving service durations, and then replay those fixed trajectories across dispatch policies.

Operational outcomes summarize ICU access, queue burden, staff utilization, and throughput \citep{Rewa2018ICUStrain}.
Admission access is measured by the admitted fraction and restricted bed-waiting time, averaged across all admission requests until bed assignment, death, or discharge before admission, or the simulation horizon.
Task queue burden measures total waiting time per initial patient across all requested tasks until service start, cancellation, or the horizon.
Task-start delay and post-ICU transfer delay are evaluated among started tasks and completed transfers, respectively.

\subsection*{Admission, care-delivery, and discharge simulations}

\paragraph{ICU bed capacity.}
For 256 test patients, we generate eight pre-ICU trajectories per patient starting from hospital admission, using each trajectory's first ICU-admission proposal as a bed request.
All patient models share these arrival requests and pre-admission trajectories, whereas each model's simulated ICU stays determine when occupied beds are released.
We evaluate first-in, first-out (FIFO) admission across 12, 16, 20, 24 and 32 ICU beds over 168 h in eight replications, measuring the admitted fraction and 168-h restricted bed-waiting time.

\paragraph{Care dispatch.}
To evaluate care-task prioritization, we compare FIFO with acute-first dispatch for 128 test patients entering simultaneously across 128 beds over 168 h.
Each patient model generates clinical trajectories and task requests with delivery feedback disabled, which are replayed under both dispatch policies across staffing levels of (8,4,8), (12,6,12) and (24,12,24) for (nurses, attending physicians, residents).
Whereas FIFO orders all pending tasks by request time, acute-first dispatch prioritizes acute care tasks over routine ones and orders tasks by request time within each priority class.
Across 16 paired replications, we estimate the acute-first-minus-FIFO contrast in acute and non-acute queue burden per initial patient.

\paragraph{Post-ICU transfer capacity.}
To evaluate downstream transfer bottlenecks, we simulate 64 test patients arriving every 2.5 h into a 24-bed unit over 168 h across eight replications under FIFO staffing of (24,4,8).
A live-exit proposal generates a downstream transfer request, and the patient continues to occupy the ICU bed until transfer service is completed.
Clinical updates and delivery feedback continue while patients await transfer, and new organ-support initiation cancels the pending transfer request.
We vary concurrent transfer slots across one, two and four with a 4-h transfer service duration, measuring post-ICU transfer delay among completed transfers and the ICU admission fraction.

\subsection*{Bed and staff capacity under arrival surges}

To examine how beds and staff jointly constrain access and care delivery, we simulate 64 test patients with pre-ICU admission proposals over 168 h.
In each run, the 64 arrivals are distributed uniformly over either 72 h or 24 h under the neural model with delivery feedback enabled and FIFO admission and dispatch.
We cross 16 or 64 ICU beds with low or high staffing of (1,1,1) and (12,4,8) for (nurses, attending physicians, residents), and also evaluate single-role expansions at 16 beds using (12,1,1), (1,4,1) or (1,1,8).
Within each of 16 paired replications, all capacity configurations share the same arrival times and patient order.
Each expansion is compared against the 16-bed, low-staff reference on restricted bed-waiting time, non-acute task-start delay, and acute and non-acute queue burden.

\subsection*{Observed-trajectory operational replay}

To test whether generated trajectories preserve the dispatch-policy effects obtained from observed clinical courses, we replay observed and model-generated ICU trajectories for 128 test patients entering a 128-bed unit simultaneously over 168 h with delivery feedback disabled.
Both trajectory sources start from each patient's observed admission support state and share the routine-care schedules, service-time distributions, and three staffing levels of the care-dispatch experiment across 16 paired replications.
To apply an identical trigger rule to both sources, acute care tasks are requested when either respiratory or circulatory support transitions from inactive to active (at most once per patient per hour, with simultaneous \textsc{stop} overriding \textsc{start} and \textsc{change} maintaining active support).
Within each source, we calculate the acute-first-minus-FIFO contrast in acute queue burden per initial patient and compare these policy contrasts as generated minus observed.

\subsection*{Matched-history analysis}

\paragraph{Cohort construction and matching.}
To test whether pre-entry clinical history carries prognostic and operational information omitted by compact state definitions, we compare patients entering single-organ support either upward from no support or downward from dual support.
On the hourly support grid, we identify transitions from no support or dual support into single-organ support (respiratory-only or circulatory-only), retaining one eligible transition per patient among test patients outside the 670-patient population-trajectory cohort.
Within exact strata of active single-organ support channel and integer hours since ICU admission ($A_t$), upward-entry and downward-entry patients are randomly paired without replacement, yielding 404 matched test pairs (808 patients).
Because matching occurs at the hourly entry boundary into single-organ support,  the dwell time of the initial state is zero ($D_t=0$) for both groups.
Each matched pair therefore supplies the Markov and semi-Markov comparators with identical initial state, hours since ICU admission and state dwell time, whereas the neural model is conditioned on each patient's observed pre-entry history without an explicit entry-direction covariate.

\paragraph{Clinical outcomes and paired comparisons.}
We follow each matched pair for 24 h on two non-mutually-exclusive endpoints: recovery to no support (reaching a state with neither respiratory nor circulatory support active at any hourly assessment from 1 to 24 h) and terminal exit (death or live ICU discharge within 24 h).
For each patient model, upward-minus-downward differences in recovery and terminal exit are estimated across 16 paired replications under unconstrained staffing (eliminating care queues) and compared against observed differences in the 404 matched pairs.

\paragraph{Mapped demand and resource use.}
Mapping \textsc{start}, \textsc{stop}, and \textsc{change} support signals to acute care tasks, we first compare upward-minus-downward differences in support-related task requests per initial patient between observed and model-generated trajectories under unconstrained staffing (\cref{fig:supp-history-task-decomposition}).
To assess how these demand differences propagate under shared resource constraints, we simulate each 404-patient group separately in a 404-bed unit with simultaneous entry over 24 h under the neural model with delivery feedback enabled.
Across staffing levels of (25,13,25), (51,25,51) and (76,38,76) for (nurses, attending physicians, residents) and 16 paired replications, we evaluate upward-minus-downward contrasts in queue burden and occupied bed time per initial patient.

\subsection*{Rollout-depth and simulation sensitivity}

To evaluate sensitivity to autoregressive rollout depth in the 670-patient cohort, we compare generated and observed trajectories across prefixes of 1 to 256 events using event-frequency total variation and transition Jensen--Shannon divergence (\cref{fig:supp-depth-precision}a).
Both metrics are regressed on $\log_2(\text{event cap})$ to estimate distributional drift per doubling of the event cap.
Across the 16 rollouts per patient, we also quantify within-patient trajectory diversity over 1 to 14 days using the proportion of unique trajectory signatures, the modal-signature fraction and mean pairwise Jaccard distance (\cref{fig:supp-depth-precision}c).

To evaluate sensitivity to simulator execution rules (\cref{fig:supp-execution-sensitivity}), we first compare 30-, 60- and 120-min clinical update intervals across four paired replications on a 128-patient cohort over 168 h, using 60 min as the reference (\cref{fig:supp-execution-sensitivity}b).
With arrivals, routine care, staffing and service rules held fixed, we compare task-start delay, proposal-to-commit latency, queue length, resource utilization, and terminal outcomes.
Finally, in the 128-patient care-dispatch experiment across all three staffing levels and 16 paired replications, we evaluate delivery-feedback sensitivity by subtracting the acute-first-minus-FIFO contrast in acute queue burden with feedback disabled from the corresponding contrast with feedback enabled (\cref{fig:supp-execution-sensitivity}c).

\subsection*{Statistical analysis}

Uncertainty from patient sampling is quantified using non-parametric bootstrap 95\% confidence intervals (2.5th to 97.5th percentiles, with curve bands showing pointwise intervals) while holding model and recalibration parameters fixed.
We use 1,000 patient-level bootstrap resamples for clinical prediction metrics, and 1,000 patient resamples paired across observed and model trajectories for the 670-patient trajectory analysis and the 256-patient model benchmark, retaining all generated rollouts for each sampled patient.
In the 404-pair matched-history analysis, 2,000 pair-level bootstrap resamples provide 95\% confidence intervals for observed clinical outcomes, observed task requests, and model clinical outcomes (after averaging the 16 unconstrained-staffing replications within each patient).
At a fixed patient cohort, finite-rollout Monte Carlo standard errors across 1 to 16 rollouts per patient are estimated from within-patient rollout variance (\cref{fig:supp-depth-precision}b).

In the discrete-event simulations, uncertainty intervals quantify Monte Carlo variability across replications conditional on the fixed evaluation cohort and model parameters.
The ICU bed-capacity experiment uses 10,000 bootstrap resamples across its eight replications, and the post-ICU transfer experiment uses two-sided 95\% Student-$t$ intervals across eight replications.
All other operational, sensitivity and matched-history contrasts (including model clinical outcomes, neural task requests, and resource-constrained outcomes) are calculated within each paired replication and summarized using two-sided 95\% paired Student-$t$ intervals across the replications specified for each experiment.

\section*{Data availability}

MIMIC-IV v3.1 is available through PhysioNet at \url{https://doi.org/10.13026/kpb9-mt58}.
Access requires PhysioNet credentialing, completion of the required research training and a signed data use agreement.

\section*{Code availability}

Our code will be provided on GitHub for academic research purposes upon publication.

\section*{Acknowledgements}

This work is supported by the National Natural Science Foundation of China under Grant 62271465, the National Key R\&D Program of China under Grant 2025YFC3408300, the Basic Research Program of Jiangsu (No. BK20255001), the Suzhou Basic Research Program under Grant SYG202338, and the Sanming Project of Medicine in Shenzhen under Grant SZSM202411033.

Yingtai Li thanks Yidan Feng (School of Nursing, The Hong Kong Polytechnic University) for her early advice on working with the MIMIC database following discussions at MICCAI 2025 in South Korea.

\section*{Author contributions}

Yingtai Li conceived and designed the study. Yingtai Li and Ruoyan Wang developed the patient world model and discrete-event simulation framework, conducted the experiments, analyzed the results and wrote the original draft. Yu Li provided domain expertise in health system simulation and health modeling. Shaohua Kevin Zhou supervised the project, provided overall guidance, and revised the manuscript. All authors read and approved the final manuscript.

\section*{Competing interests}

The authors declare no competing interests.

\bibliographystyle{unsrtnat}
\bibliography{references}

\clearpage
\begingroup

\makeatletter
\setlength{\@fptop}{0pt}
\setlength{\@fpbot}{0pt plus 1fil}
\makeatother
\graphicspath{{figures/supplement/}}

\setcounter{section}{0}
\setcounter{figure}{0}
\setcounter{table}{0}
\setcounter{equation}{0}
\renewcommand{\thesection}{S\arabic{section}}
\renewcommand{\thefigure}{S\arabic{figure}}
\renewcommand{\thetable}{S\arabic{table}}
\renewcommand{\theequation}{S\arabic{equation}}
\renewcommand{\theHsection}{supp.\arabic{section}}
\renewcommand{\theHfigure}{supp.\arabic{figure}}
\renewcommand{\theHtable}{supp.\arabic{table}}
\renewcommand{\theHequation}{supp.\arabic{equation}}
\captionsetup[figure]{name=Supplementary Figure,font={footnotesize,stretch=1}}
\crefalias{figure}{suppfigure}
\crefalias{table}{supptable}

\pdfbookmark[0]{Supplementary Information}{supplement-information}
\begin{center}
  {\Large\bfseries Supplementary Information}
\end{center}
\medskip

\noindent\begin{minipage}{\linewidth}
\captionsetup{type=figure}
\pdfbookmark[1]{Supplementary figures}{supplement-figures}
\centering
\includegraphics[width=\linewidth]{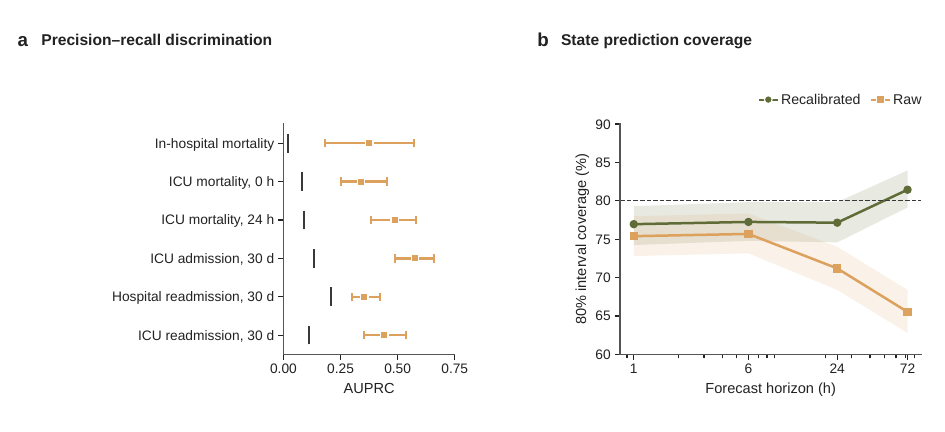}
\caption{\textbf{Precision--recall discrimination and state interval coverage.}
\textbf{a}, Area under the precision--recall curve (AUPRC) across six clinical prediction tasks, each evaluated on 1,024 records from 956--981 test patients; vertical ticks mark test-cohort task prevalence.
For ICU mortality, 0 and 24 h denote the observed history window after ICU admission.
\textbf{b}, Empirical coverage of nominal 80\% prediction intervals across 14 continuous state variables at 1, 6, 24 and 72 h, before and after validation-fitted recalibration.
Coverage was evaluated on the ten-bin state distributions using the randomized probability integral transform (PIT), measuring the proportion of observed state bins whose randomized cumulative probability fell between 0.1 and 0.9 relative to the nominal 80\% level (dashed line).
Whiskers and shaded bands show 95\% patient-cluster bootstrap intervals from 1,000 resamples.}
\label{fig:supp-calibration}
\end{minipage}
\clearpage

\begin{figure}[p]
\centering
\includegraphics[width=\linewidth]{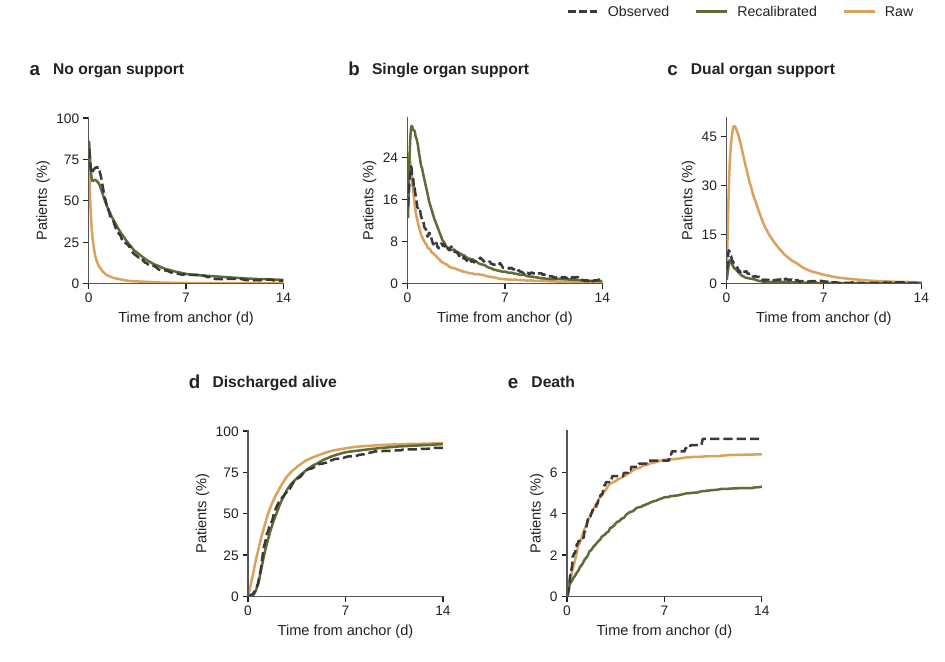}
\caption{\textbf{Hourly five-state occupancy in generated virtual cohorts.}
\textbf{a--e}, Hourly cohort proportions in no support, single-organ support, dual support, live ICU discharge and death over 14 days, comparing observed clinical courses with raw and recalibrated trajectories across 670 test patients (16 rollouts per patient per variant, averaged within each patient and weighted equally across patients).
Live discharge and death are absorbing states; vertical axis ranges differ across panels.}
\label{fig:supp-trajectory}
\end{figure}
\clearpage

\begin{figure}[p]
\centering
\includegraphics[width=\linewidth]{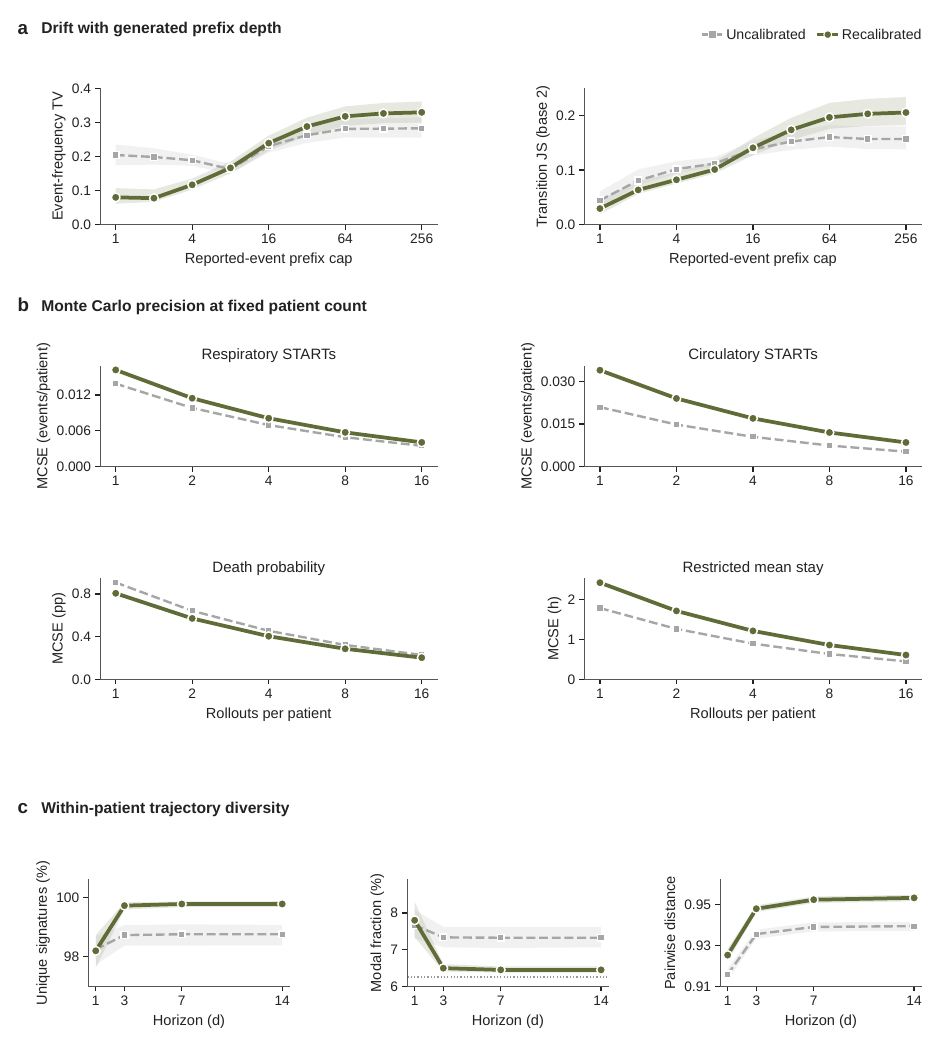}
\caption{\textbf{Generation depth, finite-rollout precision and trajectory diversity.}
\textbf{a}, Event-frequency total variation and transition Jensen--Shannon divergence across prefix caps of 1 to 256 events in the 14-day trajectories of 670 test patients (16 rollouts per patient per variant; after recalibration, drift increased by 0.0379 [95\% CI, 0.0325--0.0421] and 0.0238 [0.0197--0.0275] per cap doubling, respectively).
Shaded bands show 95\% patient-bootstrap intervals.
\textbf{b}, Estimated Monte Carlo standard errors (MCSEs) for raw and recalibrated cohort means at budgets of $R=1, 2, 4, 8$ and $16$ rollouts per patient, computed as $\widehat{\mathrm{MCSE}}(R)=\sqrt{\sum_{i=1}^{N}s_i^2/(N^2R)}$, where $s_i^2$ is the sample variance across the 16 generated rollouts for patient $i$ ($N=670$ fixed patients; at $R=16$, recalibrated MCSEs equal 33--54\% of the corresponding patient-bootstrap standard errors).
\textbf{c}, Within-patient trajectory diversity across the 16 rollouts per patient over 1 to 14 days, measured by the proportion of unique trajectory signatures, the modal-signature fraction and the mean pairwise Jaccard distance, averaged equally across patients.
Signatures comprise ordered event types, sampled event-label bins and 6-h time bins (with order and multiplicity ignored for Jaccard distance); when all 16 trajectory signatures for a patient are distinct, the modal fraction reaches its minimum of $1/16$ (6.25\%, dotted line).
Shaded bands show 95\% patient-bootstrap intervals from 1,000 resamples.
pp, percentage points.}
\label{fig:supp-depth-precision}
\label{fig:supp-trajectory-diversity}
\end{figure}
\clearpage

\begin{figure}[p]
\centering
\includegraphics[width=\linewidth]{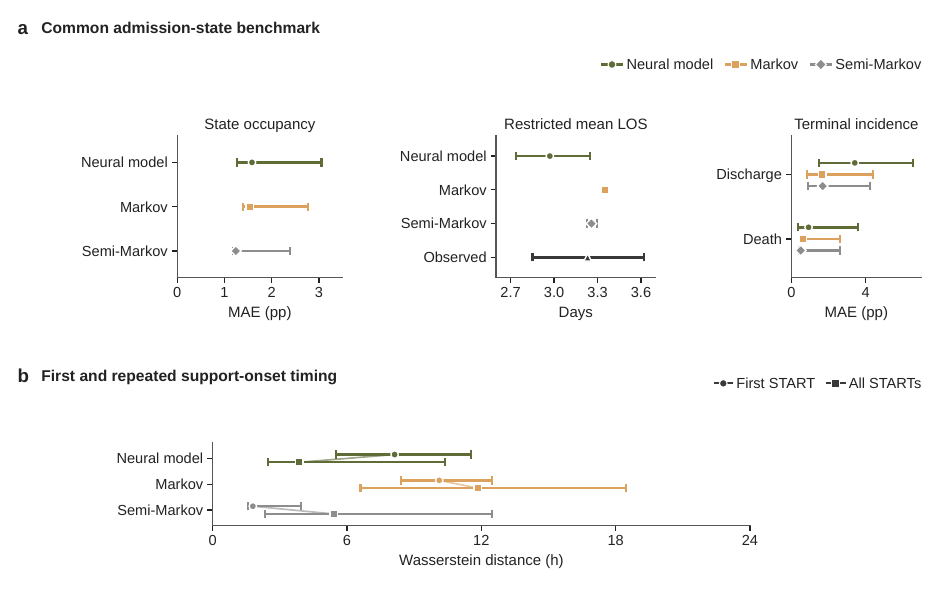}
\caption{\textbf{Markov and semi-Markov comparator benchmarks and support-onset timing.}
\textbf{a}, Cohort-level benchmark across 256 test ICU stays with matched admission-state initialization over 14 days, comparing hourly five-state occupancy mean absolute error (MAE), restricted mean ICU length of stay, and cumulative live-discharge and death incidence MAE.
Whiskers show 95\% patient-bootstrap intervals from 1,000 resamples.
\textbf{b}, First-\textsc{start} and all-\textsc{start} 1-Wasserstein distances (in hours) over 7 days for 244 eligible test patients without baseline support.
All \textsc{start}s include recurrences; marker shapes identify endpoints, lines connect estimates within each model, and whiskers show 95\% patient-bootstrap intervals.}
\label{fig:supp-engine-benchmarks}
\end{figure}
\clearpage

\begin{figure}[p]
\centering
\includegraphics[width=\linewidth]{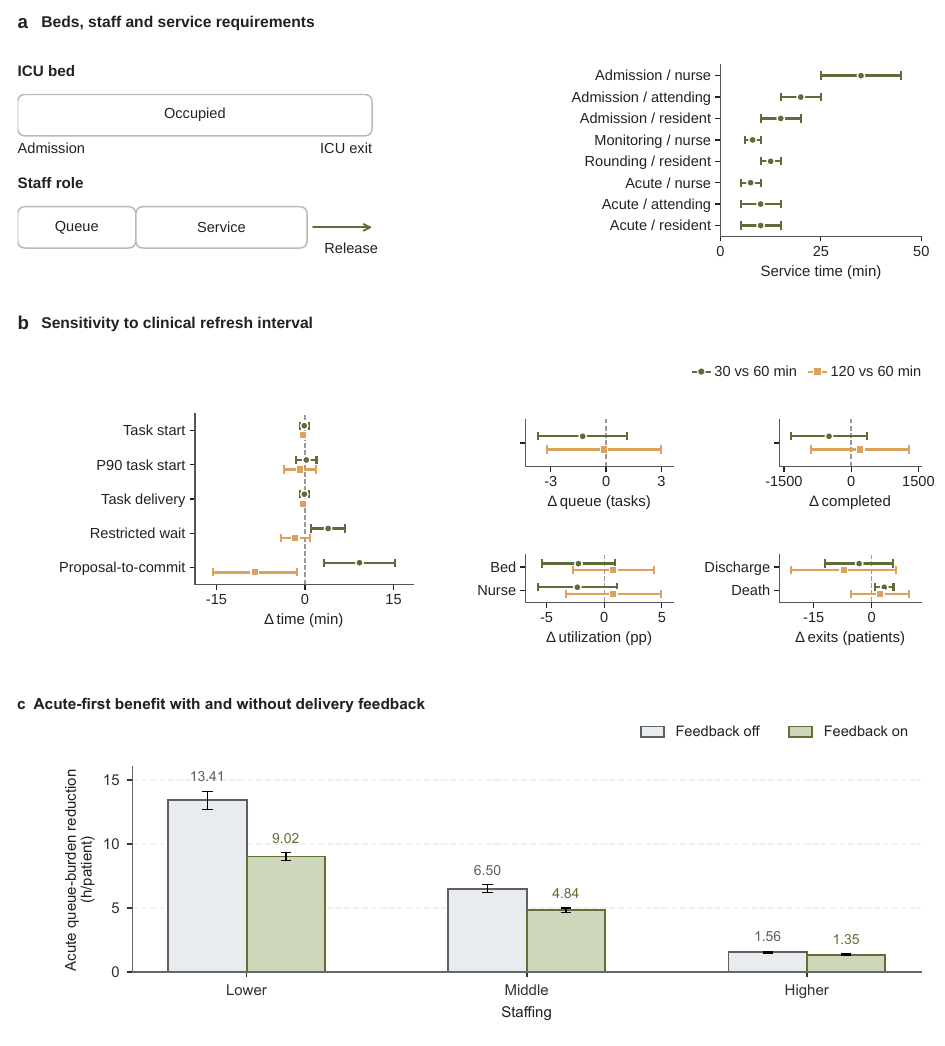}
\caption{\textbf{Resource requirements and execution sensitivity.}
\textbf{a}, Role-specific task requirements and triangular service-time distributions (minimum/mode/maximum in minutes: admission---nurse 25/35/45, attending physician 15/20/25 and resident 10/15/20; hourly monitoring---nurse 6/8/10; 12-h rounding---resident 10/12.5/15; acute care---nurse 5/7.5/10, attending physician 5/10/15 and resident 5/10/15; simulation workflow in \simulationexecutionref).
\textbf{b}, Paired 30- and 120-min clinical update-interval contrasts relative to the 60-min reference across 128 test patients over 168 h (whiskers, paired Student-$t$ 95\% intervals across four replications).
\textbf{c}, FIFO-minus-acute-first reductions in acute queue burden per initial patient with delivery feedback disabled or enabled ($128$ test patients, $128$ beds, $168\text{-h}$ horizon; bars and whiskers, means and paired Student-$t$ 95\% intervals across $16$ replications). At lower staffing, enabling feedback reduces the benefit from $13.41$ to $9.02\text{ h}$ per patient (paired attenuation $4.39\text{ h}$, 95\% interval $[3.79,5.00]$).}
\label{fig:supp-execution-sensitivity}
\end{figure}
\clearpage

\begin{figure}[p]
\centering
\includegraphics[width=\linewidth]{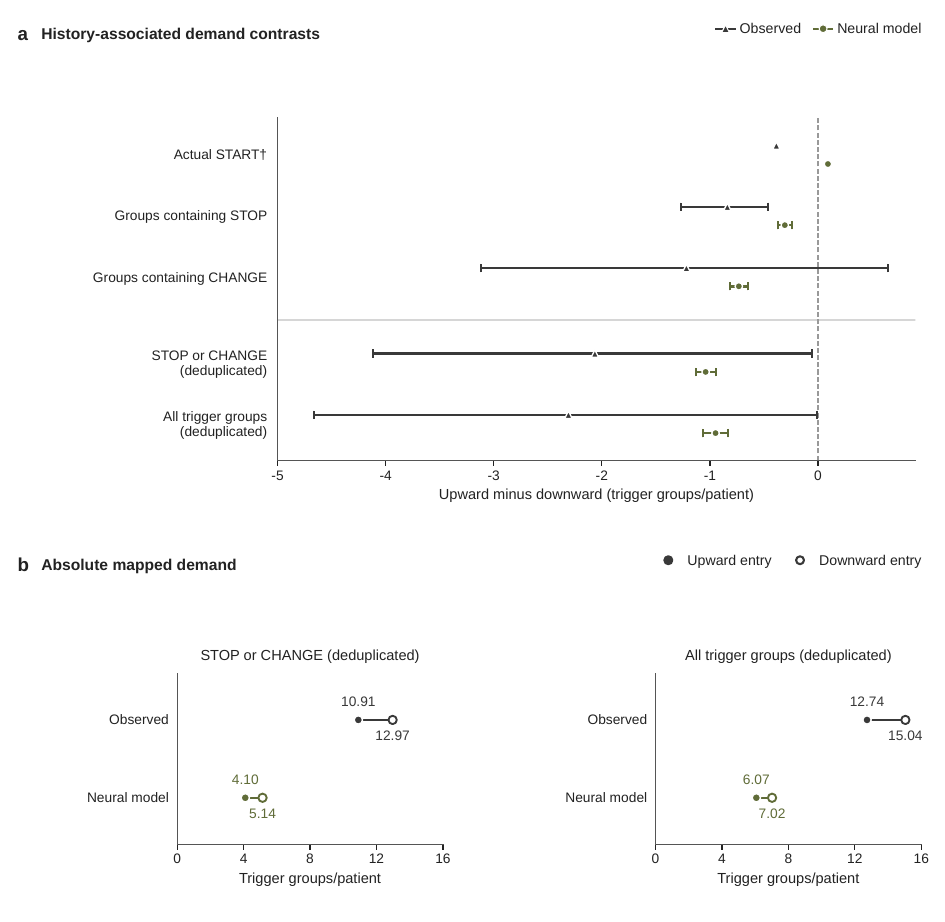}
\caption{\textbf{Organ-support triggers underlying matched-history care-demand differences.}
Values are acute trigger groups per initial patient over 24 h across 404 matched test pairs (neural estimates use 16 paired replications under unconstrained staffing).
Simultaneous support signals for a patient were grouped into a single trigger group, each generating one task request for each of the three staff roles; 14 observed groups contained multiple support-event types, so the \textsc{stop} and \textsc{change} categories overlap and deduplicated summaries count each group once.
\textbf{a}, Upward-minus-downward differences in observed and neural-generated demand by trigger group, showing that lower total acute demand in upward-entry patients was driven primarily by fewer \textsc{stop}- and \textsc{change}-related groups.
Whiskers show 95\% intervals from 2,000 pair-level bootstrap resamples for observations and paired Student-$t$ Monte Carlo (MC) intervals across the 16 replications for neural estimates; the dashed line marks zero.
$\dagger$Actual-\textsc{start} values are point estimates without confidence intervals; counting all proposed neural \textsc{start} events instead gives $1.9692$ versus $1.8773$ per patient (contrast $+0.0919$, MC 95\% CI $[0.0339,0.1499]$).
\textbf{b}, Absolute mean demand for deduplicated \textsc{stop}-or-\textsc{change} and all trigger groups in upward-entry (filled circles) and downward-entry (open circles) patients.}
\label{fig:supp-history-task-decomposition}
\end{figure}
\clearpage
\endgroup

\end{document}